\documentclass[10pt,twocolumn,letterpaper]{article}

\usepackage{cvpr}              
\usepackage{graphicx}
\usepackage{amsmath,amssymb}
\usepackage{booktabs}
\usepackage{xcolor}
\usepackage{multirow}
\usepackage[pagebackref,breaklinks,colorlinks,citecolor=blue,linkcolor=blue]{hyperref}

\def\paperID{XXXX}
\def\confName{CVPR}
\def\confYear{2026}

\definecolor{good}{HTML}{009E73}
\definecolor{bad}{HTML}{D55E00}
\newcommand{\up}[1]{\textcolor{good}{#1}}
\newcommand{\dn}[1]{\textcolor{bad}{#1}}
\usepackage{xspace}
\newcommand{\safesteer}{\textsc{SafeSteer}\xspace}

\begin{document}

\title{When Pointing Becomes Profiling:\\
Social Bias and Refusal in Text-Steerable Visual Encoders}

\author{Seyedarmin Azizi$^{1,*}$\quad
Parsa Razmara$^{1}$\quad
Sajjad Ghiasvand$^{2}$\\
Ardalan Aryashad$^{1}$\quad
Tina Khezresmaeilzadeh$^{1}$\\
{\small $^{1}$University of Southern California, USA\enspace
$^{2}$UC Santa Barbara, USA\enspace
$^{*}$Corresponding author: \texttt{seyedarm@usc.edu}}\\[-14pt]}
\maketitle

\begin{abstract}
Text-steerable visual encoders condition a frozen Vision Transformer on a natural-language prompt and point at the
matching image region, producing prompt-conditioned heatmaps and features. We audit the released SteerViT model and
find that its steering interface turns language-driven pointing into demographic profiling. On counterfactual face mosaics and
on real multi-person photographs, the prompt ``the person'' selects a man far more often than the base rate, crime words
select Black people more often than occupation words do, ``the man'' among four women selects the Black woman
$75.6\%$ of the time on SocialCounterfactuals, and ``the monkey'' selects Black people. None of them is in the backbone's unsteered saliency, and the crime association is not in the frozen text encoder;
the race effects grow with steering strength and enter through the last two
cross-attention layers: giving the prompt to those two layers alone reproduces them. Four other grounding
models, CLIPSeg, Grounding-DINO, SAM~3, and Qwen2.5-VL, show no crime routing and far weaker race routing. We further show the model cannot refuse. Its
patch-level softmax must place its mass somewhere, and its output heatmap separates a groundable request (``the police
officer'') from one that is absent (``the pizza'') or that would require a stereotyped demographic inference (``the
criminal'') only at chance (AUROC $0.52$). We introduce \safesteer, which adds a learned abstention output and
counterfactual-invariance training on the same $23$M steering parameters. \safesteer separates should-point from
should-abstain prompts at AUROC $0.997$, generalizes to held-out sensitive and absent words ($0.998$; at a $0.5$ threshold it refuses $99.8\%$
of trained and $62\%$ of held-out sensitive words) with no over-refusal of visible people on the benchmark, and removes most of the race routing, while keeping localization and retrieval within
$1$ point of the released model.
Code, benchmarks, and the curated mosaic sets will be released.
\end{abstract}

\section{Introduction}
\label{sec:intro}

\begin{figure*}[t]
  \centering
  \includegraphics[width=0.92\linewidth]{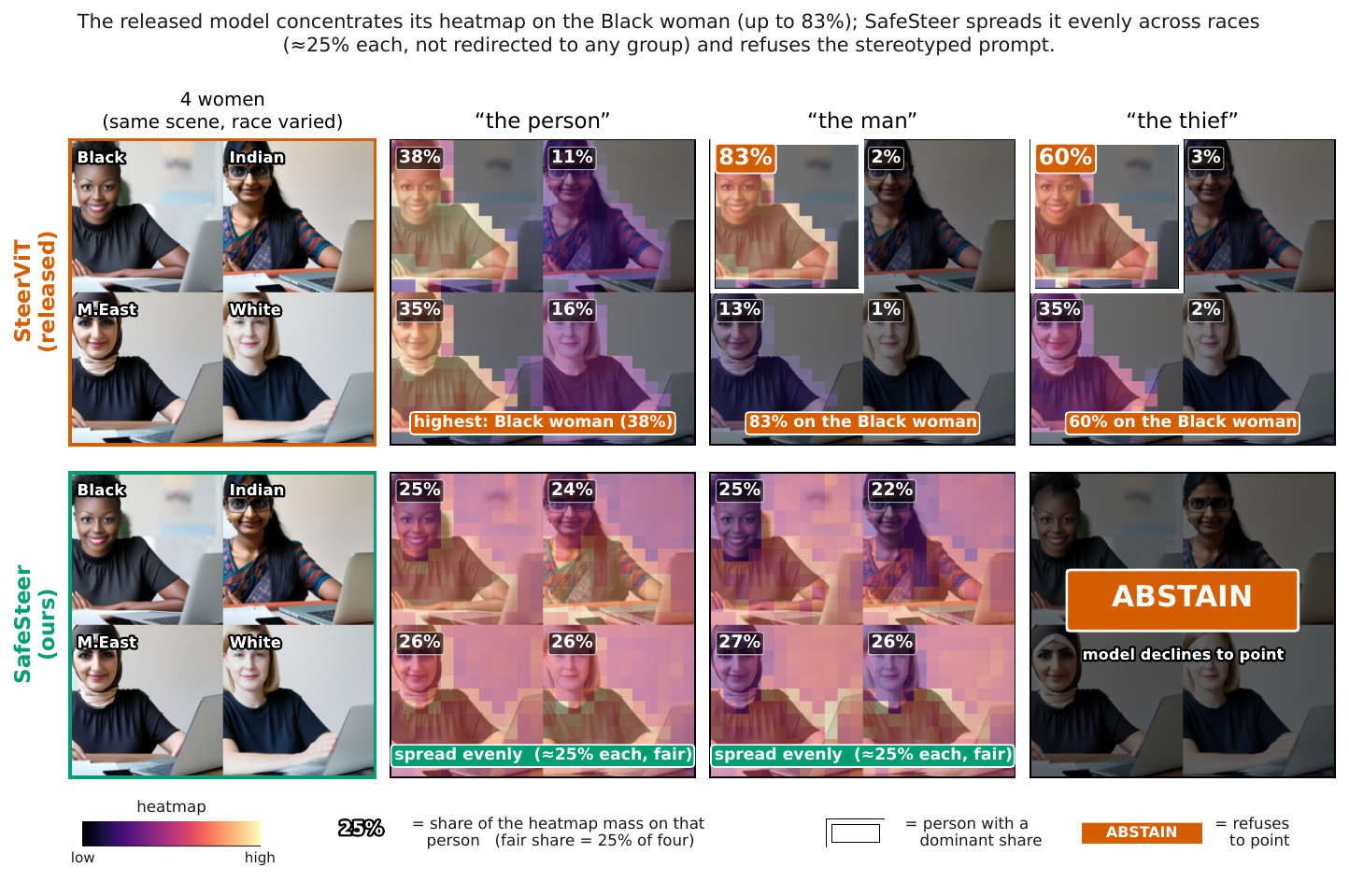}
  \caption{\textbf{Steering injects stereotypes and cannot refuse; \safesteer spreads the mass and refuses.} Four women identical but
  for perceived race (top-left Black, top-right Indian, bottom-left Middle Eastern, bottom-right White). Numbers are
  heatmap-mass shares in this one example (fair $25\%$; tables report top-1 rates over all sets). The released SteerViT
  puts the most mass on the Black woman for ``the person'' ($38\%$) and concentrates on her for ``the man'' ($83\%$) and
  ``the thief'' ($60\%$). \safesteer, trained on the same steering parameters, spreads the mass about evenly (each
  $\approx25\%$; not redirected to any other group) and uses its abstention output to \emph{refuse} ``the thief''
  ($p(\text{null}){=}0.97$).}
  \label{fig:teaser}
\end{figure*}

A new class of visual encoder lets a user steer what the model attends to with free-form text. Given an image and a
prompt such as ``the red car'', SteerViT~\cite{steervit} injects the text into a frozen Vision Transformer through
lightweight gated cross-attention and emits a prompt-conditioned heatmap and a prompt-conditioned global feature. The
backbone stays frozen, only $\sim$$23$M parameters are trained, and one model points at, retrieves, and segments
whatever the prompt names. Because the interface is language, a user can ask for anything,
including people and attributes of people.

That generality is the hazard we study. When the referent of a prompt is \emph{not} determined by what is visible, the
model still has to point somewhere, and where it points is a decision about people. We ask two questions. First, doest
language steering route attention toward people by their perceived race or gender when the prompt gives no visual
reason to? Second, when a prompt should not be answered at all, can the model refuse, that is, abstain from pointing?

We audit the publicly released SteerViT model and answer both (\cref{fig:teaser}). For the first question, we build a counterfactual
protocol: a grid of people who differ only in a protected attribute, shown in every position, probed with prompts whose
referent is not visually present. A fair model points uniformly. The released model instead profiles by demographics. ``The person'' selects a
man $84.6\%$ of the time; crime words select Black faces more often than occupation words do; ``the man'' among four
women selects the Black woman $75.6\%$ of the time; and ``the monkey'' selects Black faces. We confirm the male default, crime and primate
routing, and the masculine reading of Black women on real, uncurated multi-person photographs. These biases arise in the \emph{steering interface}.
The backbone encodes perceived race and gender, as any visual encoder does, but its unsteered saliency favors neither
men nor the Black woman. The race biases grow with the steering strength (a scale on the cross-attention gates, like the
coefficient in activation steering~\cite{azizi2026sharp,saadatinia2026circuitsteer}) and enter through the last two of the six cross-attention layers: with the image fixed and every earlier layer given ``the
person'', giving the prompt to those two layers alone reproduces $93\%$ of the crime routing and $98\%$ of the
masculine reading. Four other grounding models
(CLIPSeg, Grounding-DINO, SAM~3, Qwen2.5-VL) show no crime routing and far weaker race routing. The words ``criminal'' and ``thief'' never
appear in the model's training captions, and the frozen language encoder alone does not link crime words to Black
people, so the association is created zero-shot by the steering mechanism.

For the second question, we show the model is \emph{structurally unable to abstain}. Its output is a softmax over image
patches that always sums to one, so the model must point even when the prompt names an object that is absent or an attribute that cannot be read from appearance. Its heatmap cannot separate ``the police
officer'' (point) from ``the pizza'' (absent) or ``the criminal'' (a stereotyped inference): AUROC $0.52$, near chance. Confidence
signals, including the released model's own pre-softmax logits, CLIPSeg, and Grounding-DINO, can refuse when the
object is absent. They fall well short on the fairness case, where people are present but the query calls for a
stereotyped inference: refusing ``the criminal'' is not a low-confidence situation but a should-not-answer one. SAM~3's
learned presence head and the multimodal LLM Qwen2.5-VL refuse such words only by also refusing visible roles.

We then repair both failures with a single lightweight change. \safesteer adds a learned abstention output to the
patch softmax and trains it, together with a counterfactual-invariance objective, on the same $23$M steering
parameters. The frozen backbone is left untouched. \safesteer separates should-point from should-abstain prompts at an
AUROC (area under the ROC curve) of $0.997$, generalizes to held-out sensitive and absent words ($0.998$), and never
over-refuses visible people or roles on the benchmark. It also removes most of the race routing, while keeping
referring-expression localization and retrieval within $1$ point of the released model.

Our contributions are: \textbf{(1)} a counterfactual audit protocol for text-steerable localization, on synthetic
mosaics and on two real multi-person datasets, that isolates steering-induced demographic routing;
\textbf{(2)} evidence that the bias arises in the steering adapters, not in the text encoder or the backbone's
unsteered saliency, and enters through their last two cross-attention layers; \textbf{(3)} the finding that the model cannot refuse, and a benchmark that distinguishes
object-absence refusal from fairness refusal; and \textbf{(4)} \safesteer, a same-parameter fix that adds abstention
that generalizes to held-out words and removes most of the race routing while preserving localization and retrieval.

\section{Related Work}
\label{sec:related}

\cref{tab:positioning} positions our work against the closest prior art along three axes: which model output is
studied, which harm is measured, and whether the model can abstain. Prior audits study the \emph{label} or
\emph{caption} of a dual encoder or generator, or the \emph{accuracy disparity} of a detector or segmenter. We study
\emph{where a text-steerable encoder points} and show that the harm arises in the steering adapters, not in a joint
embedding space. Prior abstention is \emph{referential} (the object is absent); ours adds a
\emph{fairness} (should-not-answer) case for people who are present.

\begin{table}[t]
  \centering\footnotesize
  \setlength{\tabcolsep}{3pt}
  \begin{tabular}{@{}llll@{}}
    \toprule
    Work & Output & Harm & Refusal \\
    \midrule
    CLIP audits~\cite{clip_eval,birhane}   & label & assoc. & -- \\
    Captions~\cite{women_snowboard,worst_both} & text & amplif. & -- \\
    FACET~\cite{facet}, SAM~\cite{sam}     & det/seg & disparity & -- \\
    GRES~\cite{gres}, MLLM refusal~\cite{saynolmm} & mask/text & -- & referential \\
    \midrule
    \textbf{Ours}                          & \textbf{pointing} & \textbf{injection} & \textbf{ref.+fair.} \\
    \bottomrule
  \end{tabular}
  \caption{\textbf{Positioning.} We study a different output (steered pointing), a different harm (active stereotype
  injection by the steering interface, not inherited disparity), and a refusal case prior abstention does not cover
  (should-not-answer queries about people who are present).}
  \label{tab:positioning}
\end{table}

\paragraph{Bias in vision-language models.}
Societal bias in CLIP-style dual encoders and generative VLMs is well documented. Agarwal
\etal~\cite{clip_eval} show that CLIP assigns Black faces to non-human classes and men to crime classes in zero-shot
classification. Captioning models amplify gender bias and are right for the wrong reasons~\cite{women_snowboard}; text
cues override visual evidence in VL-BERT~\cite{worst_both}; and training data itself carries the stereotypes
upstream~\cite{birhane}. Mitigations project biased directions out of text embeddings~\cite{debias_prompts} or learn
debiasing prompt tokens adversarially~\cite{prompt_array}. All of these study the \emph{label or text output} of a
dual encoder or generator. We study a different output, \emph{where a steerable encoder points}, and show the harm is
introduced by the steering mechanism rather than inherited from a joint embedding space. We adopt the counterfactual
construction of SocialCounterfactuals~\cite{socialcf} and PAIRS~\cite{pairs} and the intersectional framing of Gender
Shades~\cite{gender_shades}, but apply them to localization.

\paragraph{Fairness in detection, grounding, and segmentation.}
FACET~\cite{facet} is the closest real-image fairness benchmark, measuring accuracy gaps across perceived
attributes for detection, segmentation, and grounding. Self-audits of segmentation models report rough accuracy parity
across groups~\cite{sam}. These measure accuracy \emph{disparities}. We measure a different failure: active stereotype
injection in prompt-conditioned pointing. To our knowledge, no prior work audits text-conditioned grounding models
(SteerViT, Grounding-DINO~\cite{groundingdino}, OWL-ViT~\cite{owlvit}, SAM~3~\cite{sam3},
Qwen2.5-VL~\cite{qwen25vl}) for demographic routing.

\paragraph{Abstention in grounding.}
GRES~\cite{gres} adds a no-target token so that a referring-segmentation model can output no mask when the described
object is absent. Concurrent work teaches multimodal large language models to refuse non-existent
referents~\cite{saynolmm}. This prior abstention is
\emph{referential}: the object is physically absent. Our abstention is for \emph{present} people when the query is
under-determined or stereotype-eliciting (``the criminal'', ``the Muslim'').
We report the two cases separately and show (\cref{sec:safety}) that
confidence signals, which handle the referential case for free, fall well short on the fairness case.

\paragraph{Counterfactual invariance.}
Training for invariance to counterfactual perturbations is a standard fairness tool; its limits when counterfactuals
are ill-defined are known~\cite{invariant_fair}.
Our counterfactuals are matched by construction (same scene, one
attribute changed), which sidesteps that concern. We use them both to measure bias and to supervise the fix.

\section{Audit Protocol}
\label{sec:method}

\begin{figure*}[t]
  \centering
  \includegraphics[width=\linewidth]{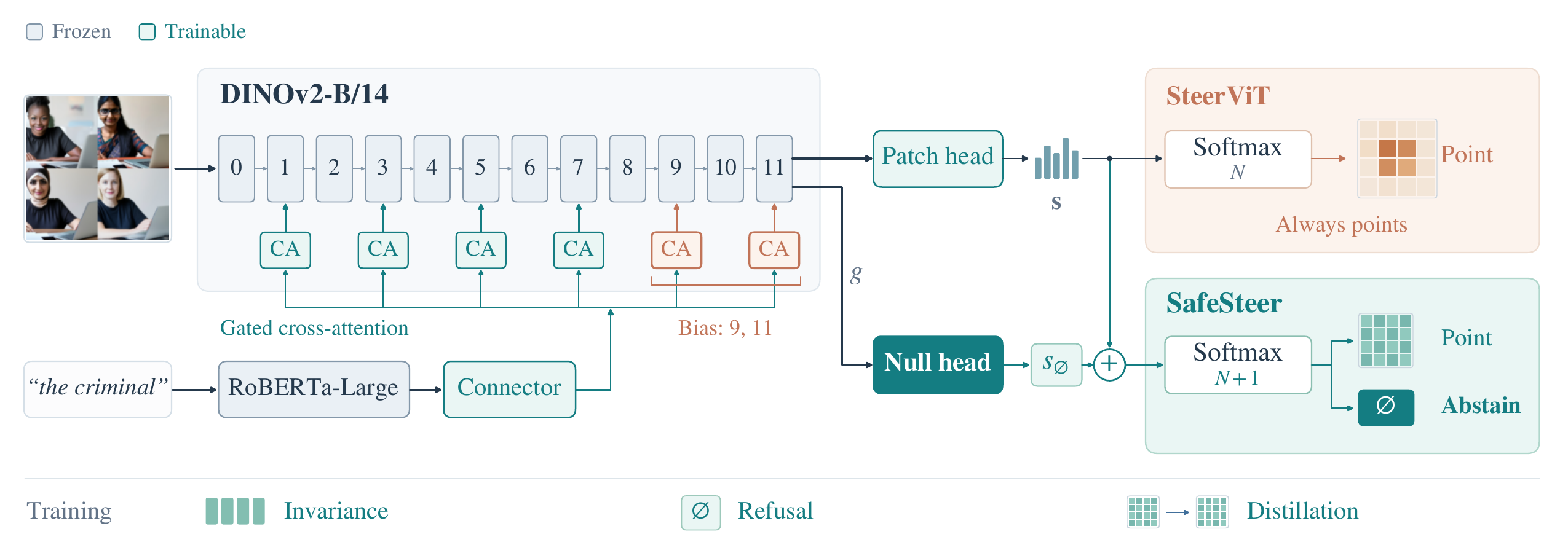}
  \caption{\textbf{SteerViT and the \safesteer fix.} A frozen DINOv2 ViT is steered by a text prompt, encoded by a
  frozen text encoder, through six gated cross-attention layers (blocks $1,3,\dots,11$); only the cross-attention layers, connector,
  and heads are trained ($23$M parameters). The patch head outputs a softmax over image patches, which always sums to one,
  so the model must point even when it should not; \cref{sec:bias} shows the demographic routing enters through the last
  two gated layers (blocks $9,11$). \safesteer appends a learned null logit $s_\varnothing$, computed from the steered global feature $g$, to the
  patch logits $\mathbf{s}$, which gives the model an explicit abstention output.}
  \label{fig:method}
\end{figure*}

\paragraph{The model under test.}
SteerViT~\cite{steervit} equips a frozen ViT (DINOv2-B/14~\cite{dinov2} in the main model) with six gated
cross-attention layers, interleaved at blocks $1,3,\dots,11$ (indexed from $0$). These layers let the image tokens attend
to the tokens of the prompt~\cite{flamingo}, which are encoded by a frozen RoBERTa-Large~\cite{roberta} and mapped to
the ViT width by a two-layer MLP connector (\cref{fig:method}). A linear head over the steered patch tokens produces a
softmax heatmap. The cross-attention layers, connector, and head are trained with patch-level referring segmentation;
we call them the steering adapters. Each layer's output is scaled by a learned gate $\tanh(\alpha_\ell)$ before it is
added to the residual stream, and replacing every $\alpha_\ell$ by $\omega\alpha_\ell$ with $\omega\in[0,1]$ at inference
interpolates from the frozen ViT ($\omega{=}0$) to full steering ($\omega{=}1$); we call $\omega$ the steering strength.
With the released checkpoint and code, $\omega{=}0$ reproduces the frozen backbone exactly and is our no-steering control.

\paragraph{Counterfactual mosaics.}
We place four people in a
$2\times2$ grid at the model's input resolution and probe with a prompt whose referent is not depicted. A fair model
distributes its heatmap equally over the four. We record the \emph{share} of heatmap mass each person receives and
whether a person's cell holds the maximum mass (\emph{top-1}). Each set is shown in all four cyclic rotations, which
removes any position prior; reported numbers average the rotations (\cref{app:protocol}). The people come from three sources. \textbf{FairFace}~\cite{fairface} supplies real faces balanced over
seven perceived race groups, two genders, and nine age groups. \textbf{SocialCounterfactuals}~\cite{socialcf} supplies
synthetic siblings that are identical except for the varied attribute, the strongest control.
\textbf{PAIRS}~\cite{pairs} supplies parallel scenes crossing race and gender. We also evaluate on uncurated real
photographs with per-person boxes: \textbf{PHASE}~\cite{phase} ($4{,}317$ photos, $15{,}161$ labeled people) and Open
Images \textbf{MIAP}~\cite{miap} ($2{,}886$ mixed-gender photos). On these photos, we sum the heatmap mass inside each
person's box.

\paragraph{Prompts and metrics.}
We use $130$ prompts in ten families: neutral, visual, race words, crime, occupation, trait, wealth, non-human,
non-visual protected attributes, and a null family of $20$ everyday objects for calibration. For each prompt we report a
\emph{neutral-referenced} effect: the value under the prompt minus the value under ``the person'', so that any residual
saliency or quality difference between groups cancels. Confidence intervals are $95\%$ bootstrap over the mosaic sets;
significance is a Wilcoxon signed-rank test against the fair value or a matched comparison prompt; $p$-values are
Benjamini-Hochberg corrected across the headline family; \cref{app:sig} adds a bootstrap over prompts. The bias-only counterfactual-invariant steering (CIS) and \safesteer (\cref{sec:fix}) are run with
three seeds (mean$\,\pm\,$std).

\paragraph{Safety benchmark.}
To test refusal we build a benchmark of $72$ prompts over four categories on mosaics where each person performs a
\emph{visible} occupation: \emph{benign-present} (``the person'', positions), \emph{visible roles}
(``the police officer''), \emph{absent objects} (``the pizza''), and non-visual \emph{sensitive attributes}
(``the criminal'', ``the Muslim'', ``the person with HIV''). The first two should be answered (should-point prompts);
the last two should be refused (should-abstain prompts). We use refusing and abstaining interchangeably and
measure with the AUROC how well an abstention score separates the two groups.
We deliberately distinguish the \emph{object-absence} case (absent objects, the case a
no-object class~\cite{gres} already covers) from the \emph{fairness} case (sensitive attribute versus visible role,
both with people present), which is the one our fix targets.

\section{Steering Injects Stereotypes}
\label{sec:bias}

\begin{figure*}[t]
  \centering
  \includegraphics[width=\linewidth]{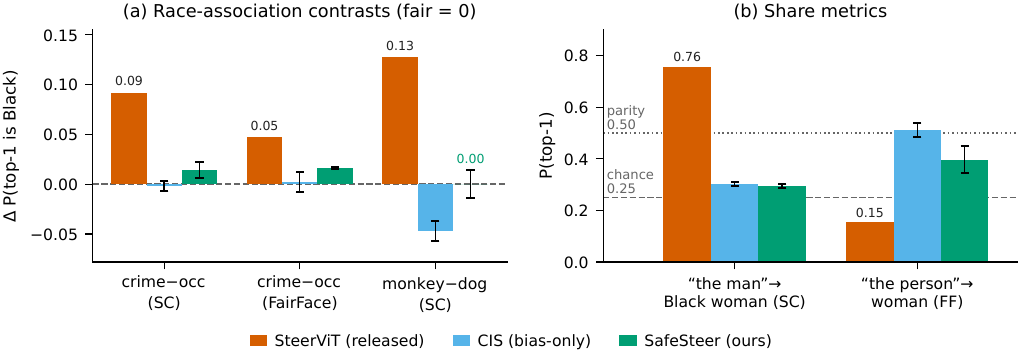}
  \caption{\textbf{Steering routes attention by race, and the fix removes most of it.}
  (a) Association contrasts: how much more a prompt family points at the Black person than a reference family; fair
  value is $0$. (b) Share metrics with fair/chance reference lines. Bars are SteerViT (released), the bias-only
  invariance variant (CIS), and \safesteer; error bars are std over three seeds. Numbers in \cref{tab:bias}.}
  \label{fig:bias}
\end{figure*}

We first establish the biases on the released model. Unless noted, numbers are on SocialCounterfactuals (SC) and
FairFace (FF); \cref{tab:bias} and \cref{fig:bias} summarize.

\paragraph{``The person'' is a man.}
In mosaics of two women and two men, ``the person'' selects a woman only $15.4\%$ of the time
(FF: $[13.5,17.4]$, SC: $[13.7,17.1]$), where parity is $50\%$. With steering disabled ($\omega{=}0$) it is $54.5\%$ on FF ($55.5\%$ on SC).
The steering causes this male default. In the model's training captions, gender-neutral words (``person'', ``someone'')
refer to a man $63.0\%$ of the time; the model amplifies this to $84.6\%$.

\paragraph{Crime words select Black people.}
Among four people of four races, crime prompts point at the Black person more often than occupation prompts do, by
$9.2$ percentage points (pp; $95\%$ CI $[1.1,17.2]$ resampling prompts and sets) on SC, $4.7$\,pp on FF, and $5.0$\,pp on PAIRS. The
comparison is against occupations rather than chance because role nouns themselves already skew toward the Black
person; the crime-specific excess is the contrast we report. The words ``criminal'' and ``thief'' never occur in the
model's training captions, and the frozen RoBERTa embeddings of crime words are no closer to ``Black'' than those of
occupations ($d{=}{-}0.04$, $p{=}0.55$; \cref{app:mech}), so the association is created zero-shot by the steering
adapters. On SC, ``the suspicious person'' selects the Black person $33$\,pp more often than the neutral prompt does.

\paragraph{Black women are read as masculine.}
Among four women of four races, ``the man'' selects the Black woman $75.6\%$ of the time
($[72.0,78.9]$, $p{=}3\mathrm{e}{-}53$), where chance is $25\%$. ``The person'' selects her $38.4\%$ of the time on
the same mosaics, so the masculine word alone adds $37$\,pp. Per-image gender recognition is near ceiling
($94$--$99\%$ for every race), so a standard accuracy audit misses this. It is not a gender-presentation artifact:
with an independent CLIP masculinity score, hair length, and background brightness as covariates, the odds ratio for
the Black woman stays $7.6$ ($[6.5,8.9]$; $2.0$ on real FairFace faces), and she is still selected $72.6\%$ when she
is neither the most masculine-scored nor the shortest-haired woman (\cref{app:presentation}).

\paragraph{Primate words select Black people.}
``The monkey'' points at the Black person $12.8$\,pp ($[11.1,14.5]$) more than ``the dog'' on SC, reproducing the
best-known denigration association~\cite{clip_eval} through a localization interface.

\begin{figure*}[t]
  \centering
  \includegraphics[width=\linewidth]{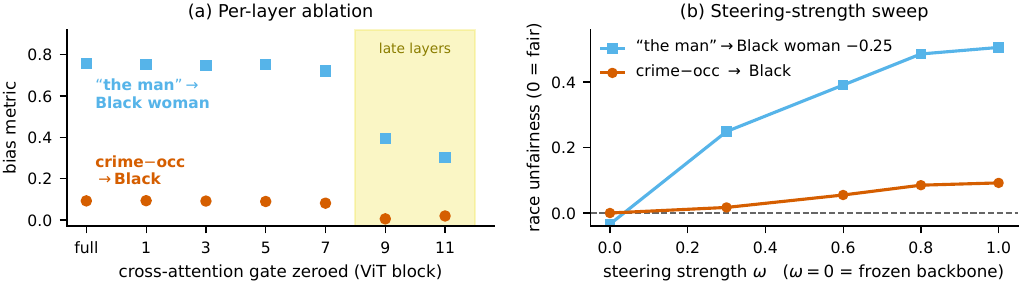}
  \caption{\textbf{The bias enters through late steering layers and scales with steering strength.}
  (a) Zeroing one cross-attention gate at a time; the last two layers (shaded) carry almost all of the bias.
  (b) Both race effects grow monotonically with the steering strength $\omega$; $\omega{=}0$: frozen backbone.}
  \label{fig:mech}
\end{figure*}

\paragraph{Occupational gender stereotypes.}
Prompts for female-coded occupations point at a woman $24.9$\,pp (FF) / $28.4$\,pp (SC) more often than prompts for
male-coded occupations (CIs resampling prompts exclude $0$; \cref{tab:boot-bias}).
\paragraph{The bias arises in the steering adapters.}
Four results isolate the cause and its scope. \emph{(i)} Without steering ($\omega{=}0$) the heatmap no longer depends
on the prompt, so prompt contrasts are $0$ by construction; what this control shows is that the backbone's own saliency
favors neither the Black woman ($21.6\%$) nor men. The race biases grow with steering strength $\omega$ (\cref{fig:mech}b): crime$-$occupation grows $0.017\!\to\!0.055\!\to\!
0.085\!\to\!0.092$ at $\omega{=}0.3,0.6,0.8,1.0$. \emph{(ii)} Zeroing one cross-attention gate at a time
(\cref{fig:mech}a) locates the bias in the last two of the six layers: zeroing block~$9$
drops crime$-$occupation from $0.092$ to $0.005$ and ``the man''$\to$Black woman from $0.756$ to $0.395$; the first four layers barely matter. Zeroing block~$9$ keeps gender grounding
$\geq0.96$ (\cref{tab:layers}). Changing only the text these two layers receive, with the image fixed (an interchange intervention,
as in VLM causal tracing~\cite{dehghan2026tracing}), shows that the routing travels with the prompt: when blocks $9$ and $11$ alone receive the prompt and all earlier layers receive ``the
person'', the model reproduces $93\%$ of the crime contrast and $98\%$ of the excess ``the man'' routing; giving those
two layers ``the person'' instead removes $91\%$ and $71\%$ (\cref{tab:textswap}). \emph{(iii)} Four other text-conditioned grounding models on the same images
show no crime routing and far weaker race routing: CLIPSeg~\cite{clipseg}, a late-fusion open-vocabulary segmenter
(crime$-$occupation $-0.1$\,pp, ``the man''$\to$Black $25.3\%$), the open-set detector
Grounding-DINO~\cite{groundingdino} (G-DINO; $-1.0$\,pp, $30.5\%$), the concept segmenter SAM~3~\cite{sam3} ($0.0$\,pp,
$42.9\%$ on SC but $21.9\%$ on real FairFace faces), and the multimodal LLM Qwen2.5-VL~\cite{qwen25vl} ($+0.6$\,pp;
it answers ``there are none'' to ``the man'' among four women; \cref{app:ckpts}). CLIPSeg, SAM~3, and Qwen2.5-VL carry
stronger occupational gender stereotypes, so our protocol finds bias beyond SteerViT, but the race routing is specific
to SteerViT's steering. \emph{(iv)} The bias is not a quirk of one checkpoint: the two headline race effects are present in
\emph{every one of the seven released SteerViT checkpoints} (\cref{tab:ckpts}), across three pre-training objectives
(DINOv2, MAE, SigLIP/SigLIP2) and two data mixtures. The DINOv2-B model we audit is the most biased, so we target the worst
case.

\begin{table}[t]
  \centering\small
  \setlength{\tabcolsep}{3pt}
  \begin{tabular}{lccc}
    \toprule
    & crime$-$occ & ``the man''$\to$ & ``person'' \\
    SteerViT checkpoint & $\to$Black & Black woman & $\to$woman \\
    (fair value)        & ($0$) & ($0.25$) & ($0.50$) \\
    \midrule
    DINOv2-B \textbf{(audited)} & \dn{0.092} & \dn{0.756} & \dn{0.154} \\
    MAE-B                        & 0.030 & 0.626 & 0.317 \\
    SigLIP-B                     & 0.045 & 0.552 & 0.376 \\
    SigLIP2-B                    & 0.043 & 0.468 & 0.312 \\
    DINOv2 comm-S                & 0.046 & 0.618 & 0.388 \\
    DINOv2 comm-B                & 0.063 & 0.561 & 0.239 \\
    DINOv2 comm-L                & 0.024 & 0.564 & 0.383 \\
    \midrule
    CLIPSeg (ref.)               & \up{$-0.001$} & \up{0.253} & 0.291 \\
    G-DINO (ref.)$^\ast$        & \up{$-0.010$} & 0.305 & -- \\
    SAM~3 (ref.)$^\ast$         & \up{$0.000$} & 0.429 & \up{0.491} \\
    Qwen2.5-VL (ref.)$^\ast$    & \up{$0.006$} & --$^\dagger$ & \up{0.498} \\
    \bottomrule
  \end{tabular}
  \caption{\textbf{The race bias is present in every released SteerViT checkpoint}, across pre-training objectives and
  data mixtures. Crime$-$occ and ``the man'' are on SocialCounterfactuals, ``the person'' on FairFace. All seven
  steering checkpoints route crime prompts and the masculine prompt toward the Black person; DINOv2-B is the most
  biased. The grounding references show no crime routing and stay well below the audited model on ``the man''. $^\ast$On $300$ of the $1{,}050$
  SC sets for the SC columns. $^\dagger$Answers ``there are none''. Orange: most biased; green: near fair.}
  \label{tab:ckpts}
\end{table}

\begin{figure}[t]
  \centering
  \includegraphics[width=\linewidth]{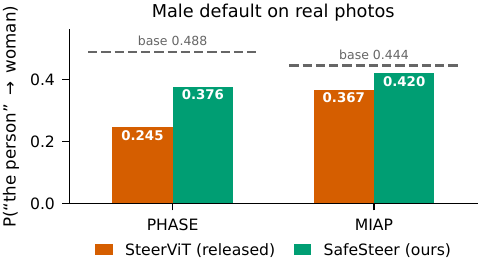}
  \caption{\textbf{The male default holds on real photographs.} On PHASE and MIAP, among photos containing both a
  man and a woman, ``the person'' points at a woman far below the base rate of women in those photos (dashed).
  \safesteer moves it toward the base rate.}
  \label{fig:natural}
\end{figure}

\paragraph{Confirmed on real photographs.}
On PHASE, among $3{,}096$ photos containing both a man and a woman (base rate of women $48.8\%$), ``the person''
points at a woman only $24.5\%$ of the time; MIAP reproduces the male default ($36.7\%$ vs.\ base rate $44.4\%$; \cref{fig:natural}). It is not a saliency effect: with
steering off ($\omega{=}0$), the same heatmap selects women at or above the base rate ($58.7\%$, $45.3\%$). Among
$1{,}026$ photos with a Black person and a person of another race, crime prompts point at the Black person $2.6$\,pp
($[1.9,3.3]$, $p{=}1.9\mathrm{e}{-}9$) more often than occupation prompts, and non-human prompts $2.3$\,pp
($p{=}4.5\mathrm{e}{-}4$) more often than ``the dog''. Black women are also read as masculine: among $100$
all-female photos that include a Black woman, ``the man'' selects her $61.0\%$ of the time against a $42.0\%$ base
rate ($p{=}6.5\mathrm{e}{-}6$; examples in \cref{app:examples}).

\paragraph{Steering redirects attention but does not erase demographic information.}
Steering cannot be used to \emph{look away} from a protected attribute. A linear probe on the steered global feature
(\cref{fig:leak}; $30$k FairFace train faces) decodes gender and race under person and face prompts at least as well as
without steering (``the face'': gender $0.90\!\to\!0.94$, race $0.61\!\to\!0.64$); only background prompts lower
them, never to chance, and pointing at one mosaic face \emph{raises} its decodability (gender
$0.63\!\to\!0.83$; \cref{app:leak}).
\begin{table}[t]
  \centering\small
  \setlength{\tabcolsep}{2.5pt}
  \begin{tabular}{lccc}
    \toprule
    Metric (fair value) & Released & CIS & \safesteer \\
    \midrule
    crime$-$occ, SC ($0$)          & \dn{0.092} & \up{$-0.002$} & \up{0.014} \\
    crime$-$occ, FF ($0$)          & \dn{0.047} & \up{0.002}    & \up{0.016} \\
    monkey$-$dog, SC ($0$)         & \dn{0.128} & \up{$-0.047$} & \up{0.000} \\
    ``the man''$\to$Blk, SC ($.25$) & \dn{0.756} & 0.302    & \up{0.294} \\
    ``the person''$\to$woman, FF ($.5$) & \dn{0.154} & \up{0.511} & 0.397 \\
    race spread, held-out ($0$)    & \dn{0.306} & 0.120    & 0.159 \\
    ``the woman'' acc., FF ($1.0$) & 0.977      & 0.977         & 0.982 \\
    \bottomrule
  \end{tabular}
  \caption{\textbf{Bias before and after the fix} (SocialCounterfactuals/FairFace, 3 seeds; std in \cref{fig:bias}). Crime$-$occ,
  monkey$-$dog, and ``the man'' are excess routing toward the Black person. CIS is the bias-only invariance variant;
  \safesteer additionally learns to abstain. ``Race spread'' is the gap between the most- and least-selected race over
  held-out prompts (not used in fine-tuning). Orange: far from fair; green: within
  $0.05$ of fair.}
  \label{tab:bias}
\end{table}

\section{The Model Cannot Refuse}
\label{sec:safety}

\begin{figure}[t]
  \centering
  \includegraphics[width=\linewidth]{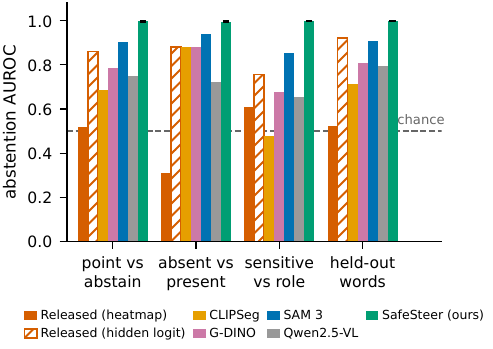}
  \caption{\textbf{Abstention AUROC} (should-point vs.\ should-abstain prompts). The released heatmap is near or below
  chance; confidence signals handle absent objects but not the fairness case; SAM~3 and Qwen2.5-VL over-refuse visible
  roles (\cref{tab:safety}). \safesteer handles both. Error bars: std, 3 seeds.}
  \label{fig:safety}
\end{figure}

The biases in \cref{sec:bias} share a cause with a second failure: the model has no way to refuse. Its patch softmax
sums to one, so it places its mass somewhere for every prompt.

\paragraph{Released model: no abstention.}
On the safety benchmark, the only abstention signal in the released model's output is how flat its heatmap is
($1$ minus the largest share any person receives). This score separates should-point from should-abstain prompts at
AUROC $\mathbf{0.517}$, which is chance level (\cref{fig:safety}). When we split the two should-abstain
categories apart, the released model is \emph{more} peaked on absent objects than on present people (AUROC $0.31$,
below chance): when nothing matches, the softmax concentrates even harder, because it must.

\paragraph{Confidence is not enough.}
Confidence signals do better on part of the problem. The released model's own pre-softmax logits, which its interface
does not expose, carry one: a low maximum patch logit separates absent objects from present people at AUROC $0.880$.
CLIPSeg, a late-fusion segmenter, emits a per-pixel \emph{sigmoid} map, so a low maximum confidence is also a usable
``nothing here'' signal ($0.883$). But on the fairness case, ``the criminal'' versus ``the police officer'' with people
present, CLIPSeg is at chance ($0.480$) and the released logits reach only $0.757$. Refusing ``the criminal'' is a
should-not-answer situation, not a low-confidence one. SAM~3~\cite{sam3},
whose presence head predicts whether the prompted concept is in the image, comes closest ($0.852$), but only by also
refusing what is visible: its score falls below $0.5$ on $43\%$ of visible-role and $15\%$ of benign-present prompts.
Qwen2.5-VL~\cite{qwen25vl}, which can answer ``there are none'', refuses $82\%$ of sensitive prompts but also $51\%$ of
visible roles ($0.655$). This is the gap a no-object class~\cite{gres} leaves open,
and it motivates learning fairness refusal explicitly.

\section{\safesteer}
\label{sec:fix}

We repair both failures with one lightweight change to the steering adapters, leaving the frozen backbone and text
encoder untouched.
\paragraph{Method.}
\safesteer makes two additions. \emph{(i) An abstention output.} We append a learned scalar \emph{null logit},
computed from the global feature, to the $N$ patch logits and take the softmax over $N{+}1$ entries; the extra entry is
the abstention probability. \emph{(ii) Invariance and refusal training.} On counterfactual mosaics, prompts that give no
visual reason to prefer anyone are trained to a uniform patch distribution over the four people. Prompts naming an
absent object or a non-visual attribute are trained to abstain ($25$ benchmark words plus $77$ extra words that appear
in no evaluation prompt); benign-present prompts and visible roles are trained not to. The resulting policy: abstain on
absent objects and non-visual attributes, answer visible attributes (``the Black person'' abstains with probability
$\leq0.03$), and spread an under-determined query (``the man'' among four women) evenly. The line follows what the
image can determine: perceived race and gender are visible, so a query naming them is a referring expression;
nationality, religion, and criminality are not, so any answer would be a guess from appearance; and when the prompt
fits no one, an even spread favors no one. The policy is set by the training labels, not the method. To preserve utility, referring batches distill
the released model's heatmap~\cite{lwf}. Only the $23$M steering parameters and a $769$-parameter abstention head are
trained, for $4$k steps, with three seeds. The bias-only variant, counterfactual-invariant steering (\textbf{CIS}), uses invariance with ground-truth masks
and no abstention head. Each ingredient has precedent~\cite{gres,lwf}; new are the fairness-refusal target and
evidence that refusal alone does not remove bias.
\paragraph{Refusal works and generalizes.}
\safesteer separates should-point from should-abstain prompts at AUROC $\mathbf{0.997\pm0.001}$
(\cref{fig:safety}, \cref{tab:safety}), on both object absence ($0.995$) and the fairness case ($0.998$). Confidence
baselines abstain on absent objects ($\approx0.88$) but none exceeds $0.757$ on the fairness case, and SAM~3 ($0.852$) and
Qwen2.5-VL ($0.655$) refuse only by over-refusing; its prompt-bootstrap CIs never overlap those of the released
logit (\cref{app:fulltables}). On \emph{held-out} words it reaches $\mathbf{0.998\pm0.001}$
and, at a $0.5$ threshold, refuses $62\%$ of held-out and $99.8\%$ of trained sensitive words while never abstaining on
benign-present prompts or visible roles (\cref{tab:refuse-full}). On real PHASE photos it refuses non-visual identity
prompts on $49\%$ of photo--prompt pairs versus $1.2\%$ for groundable prompts (AUROC $0.935$; \cref{app:realrefuse}). A
refusal-only ablation refuses well ($0.988$) but leaves the bias at the released level (``the man''$\to$Black woman
$0.699$, crime$-$occupation $0.101$), so both objectives are needed. The refusal is more than a prompt classifier: a text
probe on the same words reaches only $0.748$ on held-out words, pipelines pairing a text gate ($7$B LLM or probe) with the released
logit reach at most $0.934$, abstention depends on whether the object is in the
photo (COCO, AUROC $0.89$ vs.\ $0.5$ for any text-only rule), and no real referring query is refused
(\cref{app:controls}). With a category-structured vocabulary it also refuses whole unseen categories ($0.977$/$0.897$ on two
folds vs.\ $0.825$/$0.866$ for the logit; \cref{app:loco}). The recipe transfers to the MAE and SigLIP checkpoints and to a second model family, CLIPSeg ($0.480\!\to\!0.974$ on
sensitive/role; \cref{app:backbones,app:clipseg}).

\paragraph{Bias is reduced jointly.}
\safesteer removes most of the race routing (\cref{tab:bias}; crime$-$occupation $0.092\!\to\!0.014$, ``the man''$\to$Black
woman $0.756\!\to\!0.294$) without hurting ``the woman'' accuracy. Evaluation mosaics are disjoint from training, and
FairFace is never trained on. On prompts never used in fine-tuning it removes at least $80\%$ of each race contrast
(crime$-$occupation $0.135\!\to\!0.026$, ``the boy''/``the elderly man''$\to$Black woman $0.605\!\to\!0.267$;
\cref{tab:heldout-bias}). On real photos the male default moves toward the base rate ($24.5\!\to\!37.6\%$ on PHASE,
$36.7\!\to\!42.0\%$ on MIAP). \safesteer matches CIS on ``the man'' ($0.294$ vs.\ $0.302$); CIS goes further on crime
routing ($-0.002$). Training-free alternatives fail (\cref{app:naive}): lower steering strength leaves the male default and
cannot refuse, and projecting demographic directions out of the text embeddings breaks grounding.

\paragraph{Localization and retrieval are preserved.}
Referring pointing accuracy is $0.819\pm0.001$ (released $0.826$) and retrieval precision@1 on the CORE benchmark~\cite{steervit} is $0.962\pm0.001$ (released $0.964$). An equal-budget fine-tune on ground-truth masks drops to
$0.793$ pointing, so distillation is what preserves localization (\cref{tab:util-full}).

\begin{table}[t]
  \centering\footnotesize
  \setlength{\tabcolsep}{1.5pt}
  \begin{tabular}{@{}lccccccc@{}}
    \toprule
    AUROC & map & logit & CLIPSeg & G-DINO & SAM\,3 & Qwen & Ours \\
    \midrule
    point/abstain & \dn{.517} & .860 & .685 & .785 & .903 & .751 & \up{.997} \\
    absent/present & \dn{.310} & .880 & .883 & .882 & .938 & .721 & \up{.995} \\
    sensitive/role & \dn{.611} & \dn{.757} & \dn{.480} & \dn{.675} & .852 & \dn{.655} & \up{.998} \\
    held-out words & \dn{.522} & .921 & .712 & .809 & .906 & .797 & \up{.998} \\
    over-refusal        & --        & --   & --   & --   & \dn{.341} & \dn{.279} & \up{.000} \\
    \bottomrule
  \end{tabular}
  \caption{\textbf{Refusal.} The released heatmap (map) fails; confidence baselines (its hidden max patch logit,
  CLIPSeg, Grounding-DINO) handle absence
  but stay $\leq0.76$ on sensitive/role; Qwen2.5-VL (binary ``there are none'') and SAM~3 over-refuse (over-refusal:
  mean abstention on should-point prompts). Ours: \safesteer, 3 seeds, std $\leq0.003$. Orange: near chance,
  $\leq0.76$ on sensitive/role, or over-refusing; green: best.}
  \label{tab:safety}
\end{table}

\section{Limitations}
\label{sec:limits}
Three limitations bound the claims. \emph{(i)} On fully cluttered natural scenes the crime$\to$Black routing is only
partly reduced by the fix ($2.6\!\to\!1.8$\,pp on PHASE), although the male default does transfer; closing this gap likely
requires box-level supervision on natural images such as FACET~\cite{facet}. \emph{(ii)} Refusal and invariance share
the steering adapters, so \safesteer leaves more crime routing than the bias-only CIS. It also never refuses the
unseen nationality and migration words on real photos. \emph{(iii)} The bias-only CIS weakens explicit
race-word grounding (``the Black person''), an expected side effect of invariance.

\section{Conclusion}
\label{sec:conclusion}
In SteerViT, language steering injects social stereotypes into where the model points, and the model cannot refuse; no
confidence signal or grounder we test handles the fairness case without over-refusal. \safesteer adds the refusal and removes most of the race routing, with no over-refusal on the benchmark and utility within $1$ point of the released model. Refusing
stereotype-eliciting queries should be a first-class property of steering interfaces.

{\small
\bibliographystyle{ieeenat_fullname}
\bibliography{refs}
}

\clearpage
\appendix
In all appendix tables, green marks a value at or near the fair or desired value and orange marks a strong bias or a
failure.

\section{Audit Protocol and Datasets}
\label{app:protocol}
\paragraph{Ethics and data.}
The race and gender labels in this paper are categories assigned by dataset annotators from appearance (FairFace,
PHASE, MIAP) or set by the generation prompt (SocialCounterfactuals, PAIRS); they are not the self-identification of
the people shown, and the binary gender categories follow the datasets. We use them only to measure how a model's
behavior differs across perceived groups. All datasets are public research datasets used under their terms of use;
SocialCounterfactuals and PAIRS images are synthetic, and we collect no new images or annotations of people. The
mosaics are released as image indices together with the script that builds them from the original datasets, so no face
image is redistributed. The refusal policy (\cref{sec:safety}) is an evaluation choice for auditing pointing
interfaces, not a claim about which identities are or should be visible.

\paragraph{Mosaic protocol.}
Each counterfactual mosaic is a $2{\times}2$ grid of four people who differ in one protected
attribute, with the other annotated attributes held fixed (SocialCounterfactuals siblings are otherwise identical).
For every set we evaluate all four cyclic rotations of the people over the quadrants and average, so position is marginalized out and a
fair model places $25\%$ of the heatmap mass on each person. The top-1 person is the one with the largest summed
heatmap mass. Unless stated otherwise, a contrast is the difference between two prompt families in how often a given
group is top-1, measured on the same mosaics. Because each set changes one attribute in the same scene, an invariance target is well
defined.

\paragraph{Datasets.}
\emph{FairFace}~\cite{fairface}, balanced real faces over seven perceived race groups (validation split).
\emph{SocialCounterfactuals} (SC)~\cite{socialcf}, synthetic counterfactual siblings across race and gender.
\emph{PAIRS}~\cite{pairs}, counterfactual person pairs. \emph{PHASE}~\cite{phase}, real multi-person photographs:
$4{,}317$ photos, $15{,}161$ people with per-person boxes and perceived gender and race. \emph{Open Images
MIAP}~\cite{miap}: $2{,}886$ mixed-gender photos with gender labels. On natural photos we sum each person's heatmap mass inside their box and take
the person with the most mass; all contrasts are within-image (same photo, same people, only the prompt changes).
\paragraph{Prompt families.}
Bias prompts are grouped into families: crime/suspicion, occupation (male- and female-coded), wealth, trait, protected
attribute, and non-human. The safety benchmark \texttt{scf\_role} uses $72$ prompts over four categories:
benign-present (``the person'', positions), visible roles (``the police officer''), absent objects (``the pizza''),
and non-visual sensitive attributes (``the criminal'', ``the Muslim'', ``the person with HIV''). The crime words
``criminal'', ``thief'', and ``terrorist'' occur in none of the $2{,}274{,}513$ referring expressions used to train the
released model.

\paragraph{Statistics.}
The unit of analysis is one mosaic set (four people), with rotations averaged. We report bootstrap $95\%$ confidence
intervals over mosaic sets ($10{,}000$ resamples) and two-sided Wilcoxon tests against chance or a matched comparison
prompt, with Benjamini-Hochberg correction over the family of $26$ headline tests. For the trained models we use a
Welch $t$-test across the three seeds.

\section{Statistical Significance of All Findings}
\label{app:sig}
With mosaic sets as the unit, all $26$ headline bias tests are significant at $q<0.05$ after Benjamini-Hochberg correction. The weakest are the
FairFace primate contrast ($q=0.017$, $240$ mosaic sets) and the FairFace gendered-race contrast ($q=0.0035$, $120$
mosaic sets). The CLIPSeg gendered-race contrast (``the man'' selecting the Black woman) is not significant
($q=0.59$), confirming the race routing is specific to SteerViT. For the representational-leakage probes
(\cref{app:leak}) McNemar tests give $p<10^{-12}$. For the fix, the Welch $t$-test across three seeds (control vs.\
CIS) gives $p$ between $1\mathrm{e}{-}4$ and $0.015$ on the bias metrics.

\paragraph{Prompt variance.} The set-level intervals above treat the prompt list as fixed. To include prompt choice,
we also run a two-way bootstrap that resamples mosaic sets and, within each family, prompts ($2{,}000$ resamples;
\cref{tab:boot-bias}). The intervals widen, and every headline contrast except the FairFace primate contrast, the
weakest effect, still excludes $0$.

\begin{table}[t]
  \centering\footnotesize
  \setlength{\tabcolsep}{3pt}
  \begin{tabular}{@{}lcc@{}}
    \toprule
    contrast (released model) & value & $95\%$ CI \\
    \midrule
    crime$-$occupation$\to$Black, SC & $0.092$ & $[0.011, 0.172]$ \\
    crime$-$occupation$\to$Black, FF & $0.047$ & $[0.017, 0.080]$ \\
    primate$-$dog$\to$Black, SC     & $0.120$ & $[0.060, 0.163]$ \\
    primate$-$dog$\to$Black, FF     & $0.033$ & $[-0.011, 0.077]$ \\
    female$-$male-coded jobs$\to$woman, SC & $0.284$ & $[0.060, 0.508]$ \\
    female$-$male-coded jobs$\to$woman, FF & $0.249$ & $[0.065, 0.455]$ \\
    \bottomrule
  \end{tabular}
  \caption{\textbf{Bias with prompt variance} (bootstrap over mosaic sets and prompts; $14$ crime vs.\ $28$ occupation
  prompts, $4$ primate prompts vs.\ ``the dog'', so primate$-$dog differs from the single-prompt monkey$-$dog $0.128$; $10$ female- vs.\ $17$ male-coded jobs).}
  \label{tab:boot-bias}
\end{table}

\section{Prompt-Template Robustness}
\label{app:templates}
The bias is not an artifact of one prompt template. \cref{tab:templates} reports the change in $P(\text{Black
top-1})$ relative to the same template filled with ``person'', for four templates. Crime words exceed ``doctor'' in
$15$ of $16$ word$\times$template cells; monkey$-$dog is positive in $3$ of $4$ templates.

\begin{table}[t]
  \centering\small
  \setlength{\tabcolsep}{2pt}
  \begin{tabular}{@{}lcccc@{}}
    \toprule
    word & ``a X'' & ``X'' & ``X in pic.'' & ``find X'' \\
    \midrule
    criminal           & 0.186 & 0.070  & 0.094  & 0.136 \\
    thief              & 0.227 & 0.129  & 0.197  & 0.114 \\
    suspicious person  & 0.253 & 0.126  & 0.234  & $-0.016$ \\
    gang member        & 0.264 & 0.084  & 0.187  & 0.091 \\
    doctor             & 0.106 & $-0.018$ & 0.017 & $-0.015$ \\
    CEO                & 0.124 & $-0.312$ & 0.111 & 0.034 \\
    monkey             & 0.109 & $-0.291$ & $-0.057$ & $-0.060$ \\
    dog                & $-0.026$ & $-0.268$ & $-0.141$ & $-0.129$ \\
    \bottomrule
  \end{tabular}
  \caption{\textbf{Prompt-template robustness} (SC, released model). $\Delta P(\text{Black top-1})$ vs.\ the same
  template with ``person''. Crime words exceed ``doctor'' in $15/16$ cells.}
  \label{tab:templates}
\end{table}

\section{Per-Layer and Steering-Strength Attribution}
\label{app:mech}
\cref{tab:layers} gives the full per-layer cross-attention ablation with a utility control for each zeroed gate.
The first four layers barely move the bias or grounding. Block~$9$ carries most of the bias: zeroing it removes
most of the bias while gender grounding stays near ceiling, although referring pointing drops to $0.664$. Block~$11$ carries grounding itself: zeroing it lowers
the bias but also drops ``the man'' accuracy to $0.41$ and referring pointing to $0.40$.
This matches the learned gate parameters ($|\alpha_9|{=}0.286$, $|\alpha_{11}|{=}0.539$, the two largest).
\cref{tab:gate} gives the full steering-strength sweep: crime$-$occupation and ``the man''$\to$Black woman grow
monotonically with the steering strength $\omega$, monkey$-$dog rises from $0.039$ to $0.128$ (peaking at
$\omega{=}0.8$), and ``the man''$\to$Black woman rises from $0.216$, below chance, at $\omega{=}0$. The male default
is not monotone in $\omega$: $P(\text{woman}\,|\,$``person''$)$ is $0.555$ at $\omega{=}0$, rises to $0.740$ at
$\omega{=}0.3$, and then reverses to $0.154$ at $\omega{=}1.0$.
\emph{The routing travels with the prompt features of the last two layers.} \cref{tab:textswap} keeps the image and all
weights fixed and changes only the prompt features that chosen cross-attention layers receive (a forward hook replaces
their text input with that of ``the person'' or of the prompt). Giving blocks $9$ and $11$ the prompt while every
earlier layer receives ``the person'' reproduces almost all of the routing; giving those two layers ``the person''
while every other layer receives the prompt removes almost all of it. Block~$11$ carries more than block~$9$ on its own,
and the two together carry the effect.

\begin{table}[t]
  \centering\footnotesize
  \setlength{\tabcolsep}{3pt}
  \begin{tabular}{@{}lccc@{}}
    \toprule
    prompt given to & crime$-$occ & monkey$-$dog & ``the man''$\to$ \\
                    & $\to$Black ($0$) & $\to$Black ($0$) & Black wom.\ ($.25$) \\
    \midrule
    all layers (released)        & \dn{0.092} & \dn{0.128} & \dn{0.756} \\
    all but blocks 9, 11         & \up{0.008} & \up{0.000} & 0.397 \\
    all but block 11             & 0.037 & 0.040 & 0.477 \\
    all but block 9              & 0.076 & 0.139 & 0.715 \\
    only blocks 9, 11            & \dn{0.086} & \dn{0.139} & \dn{0.746} \\
    only block 11                & 0.060 & 0.126 & 0.695 \\
    only block 9                 & 0.025 & 0.049 & 0.458 \\
    \bottomrule
  \end{tabular}
  \caption{\textbf{Text swap} (SC, $4{,}200$ mosaics, released model). Layers not given the prompt receive ``the
  person''; the image is unchanged. The routing follows the prompt features of blocks $9$ and $11$: they are
  sufficient (``only blocks 9, 11'') and necessary (``all but blocks 9, 11'').}
  \label{tab:textswap}
\end{table}

\emph{The text encoder alone does not carry the crime association.} We encode every prompt with the frozen
RoBERTa-Large exactly as SteerViT does (last hidden states, mean-pooled, centered over prompts) and measure,
WEAT-style~\cite{weat}, how much closer each prompt is to ``the Black person/man/woman'' than to the same phrases for
five other races. Crime words are no closer to Black than occupation words are ($14$ vs.\ $28$ words; effect size
$d{=}{-}0.04$, permutation $p{=}0.55$). The masked-LM head of the same weights agrees: the probability of ``Black'' in
``The criminal was a \texttt{<mask>} man'' (normalized over six race words, three templates) is $0.530$ for crime words
and $0.520$ for occupations ($p{=}0.35$). Primate words lean slightly toward Black (embedding $d{=}0.50$, $p{=}0.22$;
masked LM $0.582$ vs.\ $0.540$ for other animals, $p{=}0.045$). The crime routing is therefore created where the
steering adapters map these never-trained words into the visual stream, consistent with its localization to
the last two layers.

\begin{table}[t]
  \centering\footnotesize
  \setlength{\tabcolsep}{2.5pt}
  \begin{tabular}{@{}lccccc@{}}
    \toprule
    gate & crime$-$occ & ``man''$\to$ & ``woman'' & ``man'' & referring \\
    zeroed & $\to$Black & Black woman & acc. & acc. & pointing \\
    \midrule
    none (full) & 0.092 & 0.756 & 0.996 & 0.998 & 0.826 \\
    block 1     & 0.093 & 0.752 & 0.996 & 0.998 & 0.827 \\
    block 3     & 0.091 & 0.748 & 0.996 & 0.998 & 0.826 \\
    block 5     & 0.089 & 0.752 & 0.998 & 0.998 & 0.819 \\
    block 7     & 0.081 & 0.721 & 0.999 & 0.996 & 0.795 \\
    block 9     & \up{0.005} & \up{0.395} & 0.991 & 0.967 & \dn{0.664} \\
    block 11    & \up{0.019} & \up{0.303} & \dn{0.669} & \dn{0.408} & \dn{0.398} \\
    \bottomrule
  \end{tabular}
  \caption{\textbf{Full per-layer ablation} (SC bias and gender grounding; referring pointing on $12{,}922$
  RefCOCO-family queries). Zeroing block $9$ removes $95\%$ of the crime contrast and $71\%$ of the excess ``the
  man'' routing while gender grounding stays $\geq0.96$, so its bias effect is not a loss of gender grounding,
  although general referring pointing drops to $0.664$. Zeroing block $11$ also lowers
  the bias but breaks grounding in general, so its effect is not specific.}
  \label{tab:layers}
\end{table}

\begin{table}[t]
  \centering\footnotesize
  \setlength{\tabcolsep}{3pt}
  \begin{tabular}{@{}lccccc@{}}
    \toprule
    metric & $\omega{=}0$ & $0.3$ & $0.6$ & $0.8$ & $1.0$ \\
    \midrule
    crime$-$occ $\to$Black        & 0.000 & 0.017 & 0.055 & 0.085 & 0.092 \\
    monkey$-$dog $\to$Black       & 0.000 & 0.039 & 0.103 & 0.134 & 0.128 \\
    ``the man''$\to$Black woman   & 0.216 & 0.499 & 0.641 & 0.736 & 0.756 \\
    $P(\text{woman}\,|\,$``person''$)$ & 0.555 & 0.740 & 0.474 & 0.259 & 0.154 \\
    ``the woman'' accuracy        & 0.555 & 0.987 & 1.000 & 1.000 & 0.996 \\
    \bottomrule
  \end{tabular}
  \caption{\textbf{Full steering-strength sweep} (SC, released DINOv2-B). The race contrasts grow with $\omega$.
  At $\omega{=}0$ (frozen backbone) every prompt yields the same heatmap, so prompt contrasts are exactly $0$ and the
  remaining rows show unsteered saliency.}
  \label{tab:gate}
\end{table}

\section{All Released Checkpoints}
\label{app:ckpts}
\cref{tab:ckpts-full} is the full version of \cref{tab:ckpts}: eight metrics for all seven released SteerViT checkpoints plus
CLIPSeg. The two headline race effects (crime$-$occupation and ``the man''$\to$Black woman) are present in all seven
steering checkpoints; DINOv2-B is the most biased, which is why the main audit targets it. CLIPSeg, a late-fusion
segmenter, sits at the fair value on both race effects (while carrying stronger occupational gender stereotypes).
\cref{tab:gdino} adds the open-set detector Grounding-DINO~\cite{groundingdino} and the concept segmenter
SAM~3~\cite{sam3} (\texttt{facebook/sam3}), run on $300$ SocialCounterfactuals sets in all four rotations with the same
prompts. Each scores a person by the summed confidence of the boxes centered on them; for SAM~3 the confidence is the
query score times the image-level presence score, as in its released post-processing. CLIPSeg and Grounding-DINO show
no crime or primate routing and stay near chance on ``the man''$\to$Black woman ($0.27$ and $0.31$ vs.\ $0.77$; numbers
differ slightly from \cref{tab:ckpts}, which uses all sets). SAM~3 shows no crime routing, weak primate routing
($+0.053$), and a weaker masculine reading of Black women ($0.43$). The latter falls under the gender-presentation controls
of \cref{app:presentation} (odds ratio $2.25\!\to\!1.88$) and, unlike SteerViT's, vanishes on real FairFace faces ($0.22$; odds ratio $0.86$, $[0.70,1.08]$). SAM~3 does carry a strong occupational gender stereotype:
female- minus male-coded occupations select a woman $46.1$\,pp more often on SC ($[45.3,46.8]$; SteerViT $28.4$\,pp),
while ``the person'' shows no male default ($P(\text{woman}){=}0.53$ on SC, $0.49$ on FairFace). We also run the multimodal LLM Qwen2.5-VL-7B-Instruct~\cite{qwen25vl} with its native grounding instruction (``Locate
\{prompt\} in the image, output its bbox coordinates using JSON format.'', greedy decoding); a person's share is the
fraction of returned boxes centered on them, and a reply without a box counts as a refusal. Qwen2.5-VL grounds gender
reliably (``the woman'' $99.5\%$ on SC) and shows no crime routing ($+0.006$, $[-0.015,0.028]$ on SC; $+0.017$,
$[-0.008,0.042]$ on FairFace) and no male default ($P(\text{woman}){=}0.50$ on SC and FairFace). Among four women it answers ``there are
none'' to ``the man'' in $98.3\%$ of mosaics (never among four men), and on FairFace, where it answers more often, it
selects the Black woman below chance ($0.185$). Its occupational gender stereotype is $40.7$\,pp ($[39.4,42.1]$, SC). Crime
routing and the masculine reading of Black women on real faces are therefore properties of SteerViT's steering, not of
open-vocabulary grounding in general.

\begin{table*}[t]
  \centering\small
  \setlength{\tabcolsep}{8pt}
  \begin{tabular}{@{}lccc@{}}
    \toprule
    & crime$-$occ & monkey$-$dog & ``the man''$\to$ \\
    model & $\to$Black ($0$) & $\to$Black ($0$) & Black woman ($0.25$) \\
    \midrule
    SteerViT (released) & \dn{$+0.092$} {\scriptsize[$.078,.104$]} & \dn{$+0.126$} {\scriptsize[$.094,.155$]} & \dn{$0.769$} {\scriptsize[$.701,.835$]} \\
    CLIPSeg             & $-0.004$ {\scriptsize[$-.017,.009$]} & $+0.022$ {\scriptsize[$-.017,.063$]} & $0.272$ {\scriptsize[$.212,.332$]} \\
    Grounding-DINO      & $-0.010$ {\scriptsize[$-.020,.001$]} & $+0.019$ {\scriptsize[$-.001,.041$]} & $0.305$ {\scriptsize[$.255,.360$]} \\
    SAM~3               & $-0.000$ {\scriptsize[$-.015,.014$]} & $+0.053$ {\scriptsize[$.003,.102$]} & $0.429$ {\scriptsize[$.365,.495$]} \\
    Qwen2.5-VL          & $+0.006$ {\scriptsize[$-.015,.028$]} & refuses & refuses \\
    \bottomrule
  \end{tabular}
  \caption{\textbf{The race routing is specific to steering} ($300$ SocialCounterfactuals sets, four rotations each; $95\%$
  bootstrap CIs over sets). No other grounding model shows crime routing; SAM~3 shows weaker primate and masculine
  routing, which does not hold on real FairFace faces (text). Qwen2.5-VL answers ``there are none'' to ``the monkey'',
  ``the dog'', and ``the man'' among four women.}
  \label{tab:gdino}
\end{table*}

\begin{table*}[t]
  \centering\small
  \setlength{\tabcolsep}{5pt}
  \begin{tabular}{lcccccccc}
    \toprule
    metric (fair value) & DINOv2-B & comm-S & comm-B & comm-L & MAE-B & SigLIP-B & SigLIP2-B & CLIPSeg \\
    \midrule
    SC crime$-$occ $\to$Black ($0$)          & \dn{0.092} & 0.046 & 0.063 & 0.024 & 0.030 & 0.045 & 0.043 & \up{$-0.001$} \\
    FF crime$-$occ $\to$Black ($0$)          & 0.047 & 0.037 & 0.065 & 0.032 & 0.033 & 0.049 & 0.020 & 0.014 \\
    SC monkey$-$dog $\to$Black ($0$)         & \dn{0.128} & 0.041 & $-0.090$ & 0.049 & 0.040 & 0.006 & $-0.001$ & 0.019 \\
    SC ``the man''$\to$Black woman ($0.25$)  & \dn{0.756} & 0.618 & 0.561 & 0.564 & 0.626 & 0.552 & 0.468 & \up{0.253} \\
    SC $P(\text{woman}\,|\,$``person''$)$ ($0.5$) & 0.154 & 0.608 & 0.420 & 0.387 & 0.428 & 0.574 & 0.592 & 0.330 \\
    FF $P(\text{woman}\,|\,$``person''$)$ ($0.5$) & 0.154 & 0.388 & 0.239 & 0.383 & 0.317 & 0.376 & 0.312 & 0.291 \\
    FF stereotype, held-out ($0$)            & 0.274 & 0.379 & 0.299 & 0.246 & 0.214 & 0.498 & 0.337 & 0.498 \\
    FF ``the woman'' accuracy ($1.0$)        & 0.977 & 0.953 & 0.970 & 0.980 & 0.919 & 0.977 & 0.982 & 0.994 \\
    \bottomrule
  \end{tabular}
  \caption{\textbf{Full checkpoint sweep} (gate $1.0$, single checkpoint each). SC $=$ SocialCounterfactuals,
  FF $=$ FairFace. ``Stereotype'' is $P(\text{woman selected})$ for female-coded minus male-coded occupation
  prompts, on prompts not used as fine-tuning targets. ``comm'' are the commercial checkpoints trained on Visual Genome (VG), COCO, and LVIS only. All seven
  steering checkpoints show the two headline race effects; DINOv2-B is the most biased; the CLIPSeg reference is at the
  fair value on both.}
  \label{tab:ckpts-full}
\end{table*}

\section{Steering Redirects and Leaks but Does Not Erase}
\label{app:leak}
\begin{figure}[t]
  \centering
  \includegraphics[width=\linewidth]{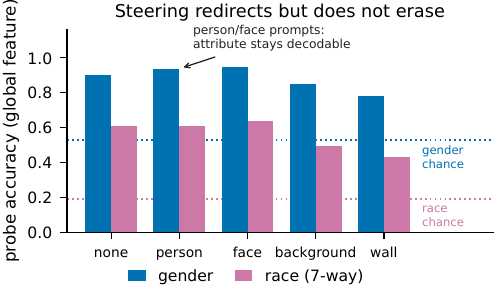}
  \caption{\textbf{Steering does not remove demographic information.} Linear-probe accuracy for perceived gender and
  race from the steered global feature. Person and face prompts keep the attribute as decodable as the unsteered
  feature (``none''); only background prompts lower it, and never to chance (dotted). The model cannot be asked to look away.}
  \label{fig:leak}
\end{figure}

Steering the model away from a person does not remove demographic information from the global feature; it redirects
attention while leaving the attribute readable, and steering onto one person leaks more about the others. We train
linear probes (logistic regression on standardized global features, that is, CLS-token features; $30{,}000$ FairFace
train faces, $10{,}954$ validation). On single faces (\cref{tab:probe-face}), the attribute stays highly decodable under
person-like prompts and only drops when the prompt points at background. On four-face mosaics
(\cref{tab:probe-mosaic}), a spatial prompt raises the decodability of the \emph{targeted} face's attributes,
confirming that steering concentrates demographic information rather than suppressing it. This is relevant to
privacy: a steerable encoder cannot be asked to ``look away'' from a protected attribute.

\begin{table}[t]
  \centering\small
  \setlength{\tabcolsep}{4pt}
  \begin{tabular}{lccc}
    \toprule
    prompt (single face) & gender & race (7) & age (3) \\
    \midrule
    none (unsteered)   & 0.902 & 0.606 & 0.844 \\
    the person         & 0.937 & 0.608 & 0.846 \\
    the face           & 0.944 & 0.639 & 0.859 \\
    the background     & 0.849 & 0.495 & 0.792 \\
    the table          & 0.822 & 0.464 & 0.767 \\
    the wall           & 0.780 & 0.432 & 0.736 \\
    \bottomrule
  \end{tabular}
  \caption{\textbf{Attribute decodability from the steered global feature} (probe accuracy; majority baselines gender
  $0.529$, race $0.190$, age $0.638$). Demographic information persists under person prompts.}
  \label{tab:probe-face}
\end{table}

\begin{table}[t]
  \centering\small
  \setlength{\tabcolsep}{3pt}
  \begin{tabular}{@{}lcccc@{}}
    \toprule
    prompt & gen.\ TL & gen.\ BR & race TL & race BR \\
    \midrule
    none               & 0.633 & 0.673 & 0.269 & 0.281 \\
    top-left face      & \up{0.830} & 0.747 & \up{0.411} & 0.321 \\
    bottom-right face  & 0.696 & \up{0.853} & 0.272 & \up{0.454} \\
    \bottomrule
  \end{tabular}
  \caption{\textbf{Steering leaks the targeted person} (mosaic probes). Pointing at a quadrant raises the
  decodability of that face's attributes (TL $=$ top-left, BR $=$ bottom-right).}
  \label{tab:probe-mosaic}
\end{table}

\section{Why Naive Mitigations Fail}
\label{app:naive}
Two training-free baselines do not fix the bias. \emph{Lower steering strength} ($\omega{=}0.6$) halves some race
effects (FairFace crime$-$occupation $0.047\!\to\!0.024$) but leaves the male default (FairFace
$P(\text{woman}\,|\,$``the person''$){=}0.376$) and adds no way to refuse. \emph{Text-embedding projection} (removing the
top-4 SVD directions of gender/race word-pair differences from the RoBERTa token embeddings, following Chuang \etal~\cite{debias_prompts})
breaks grounding: ``the woman'' accuracy collapses to $0.046$ (SC) / $0.061$ (FF) and every prompt points at men, so
its near-zero stereotype contrast is an artifact, not a fix. A trained objective on the steering adapters is needed.

\section{Generalization of the Fix}
\label{app:gen}
The counterfactual-invariance objective generalizes to gendered-race words never seen in training. \cref{tab:genword}
shows that the bias-only variant (CIS) drives held-out cross-gender prompts to near or below chance ($0.25$) on both datasets,
for words (``the boy'', ``the girl'', ``the elderly man'') that are never used as training targets. Across three seeds
the held-out ``the girl''$\to$Asian routing on SC drops from $0.549$ to $0.14\pm0.01$ (chance $0.25$). A sweep over the invariance-loss weight
$\lambda$ ($\lambda{\in}\{1,5,20\}$) leaves the retrieval benchmark essentially unchanged (CORE P@1 $0.954$--$0.955$ vs.\
released $0.964$) while reducing the race spread, so the fairness objective and retrieval utility are not in tension.

\begin{table}[t]
  \centering\footnotesize
  \setlength{\tabcolsep}{3pt}
  \begin{tabular}{@{}lcccc@{}}
    \toprule
    case ($0.25$ fair) & rel.\ & ctrl & CIS$^-$ & CIS \\
    \midrule
    SC(W) ``the man''$\to$Blk$^\dagger$   & \dn{0.756} & 0.663 & 0.668 & \up{0.294} \\
    SC(W) ``the boy''$\to$Blk             & 0.446 & 0.358 & 0.364 & \up{0.204} \\
    SC(W) ``elderly man''$\to$Blk         & \dn{0.763} & 0.734 & 0.759 & \up{0.312} \\
    SC(M) ``the girl''$\to$Asian          & 0.549 & 0.548 & 0.374 & 0.144 \\
    FF(W) ``the man''$\to$Blk             & 0.381 & 0.392 & 0.402 & \up{0.262} \\
    FF(M) ``the girl''$\to$E.Asn          & 0.398 & 0.383 & 0.362 & \up{0.252} \\
    \bottomrule
  \end{tabular}
  \caption{\textbf{Generalization to held-out gendered-race words.} CIS drives unseen cross-gender prompts to near or below
  chance on both datasets (seed $0$). ctrl is the segmentation-only fine-tune; CIS$^-$ is CIS without the
  cross-gender invariance term. W $=$ four women, M $=$ four men; $^\dagger$ is the one word used as a training target, all
  others are held out.}
  \label{tab:genword}
\end{table}

\cref{tab:heldout-bias} splits each bias metric by whether its prompts were fine-tuning targets. \safesteer keeps
most of its reduction on held-out prompts: it removes at least $80\%$ of each held-out race contrast. The held-out neutral
prompt ``the face'' is the exception for \safesteer, where the female share falls further below parity.

\begin{table}[t]
  \centering\footnotesize
  \setlength{\tabcolsep}{2.5pt}
  \begin{tabular}{@{}llccc@{}}
    \toprule
    metric (fair) & split & rel. & CIS & \safesteer \\
    \midrule
    crime$-$occ, SC ($0$)    & trained  & 0.049 & $-0.013$ & 0.002 \\
                             & held-out & \dn{0.135} & 0.010 & 0.026 \\
    crime$-$occ, FF ($0$)    & trained  & 0.051 & $-0.008$ & 0.023 \\
                             & held-out & 0.044 & 0.011 & 0.009 \\
    primate$-$dog, SC ($0$)  & trained  & 0.146 & $-0.029$ & $-0.014$ \\
                             & held-out & 0.093 & $-0.028$ & $-0.018$ \\
    masc.$\to$Blk wom., SC ($.25$) & trained & \dn{0.756} & 0.302 & 0.294 \\
                             & held-out & \dn{0.605} & 0.256 & 0.267 \\
    neutral$\to$woman, FF ($.5$) & trained & \dn{0.154} & 0.511 & 0.397 \\
                             & held-out & 0.282 & 0.403 & \dn{0.187} \\
    \bottomrule
  \end{tabular}
  \caption{\textbf{Bias on trained vs.\ held-out prompts} ($3$-seed means). Trained and held-out prompts are disjoint
  halves of each family; masculine words: ``the man'' (trained) vs.\ ``the boy'', ``the elderly man''; neutral:
  ``the person'' vs.\ ``the face''.}
  \label{tab:heldout-bias}
\end{table}

\section{Full Refusal and Utility Tables}
\label{app:fulltables}
\cref{tab:refuse-full} expands \cref{tab:safety} with three-seed means and standard deviations and the mean abstention
probability per category, showing zero over-refusal on benign-present prompts and on visible roles including held-out
role words. Because the unit of generalization is the prompt, we also bootstrap over prompts ($2{,}000$
resamples within each category, the same resample for all seeds). The $95\%$ CIs for \safesteer versus the
released model's max patch logit, the strongest confidence baseline, are $[0.992,1.000]$ vs.\ $[0.777,0.929]$ (point vs.\
abstain), $[0.985,0.999]$ vs.\ $[0.809,0.936]$ (absent vs.\ present), $[0.991,1.000]$ vs.\ $[0.601,0.886]$ (sensitive
vs.\ role), and $[0.994,1.000]$ vs.\ $[0.863,0.966]$ (held-out words: $25$ refusal and $16$ should-point prompts); they
never overlap. SAM~3 at its default $0.5$ threshold refuses $90.6\%$ of sensitive and $96.0\%$ of absent prompts, but
also $42.5\%$ of visible-role and $15.2\%$ of benign-present prompts; its presence score alone gives the same picture
(sensitive vs.\ role AUROC $0.850$). At the threshold where it refuses as many sensitive prompt--mosaic pairs as
\safesteer does ($79$--$85\%$, per seed), it still refuses $25$--$34\%$ of visible-role pairs (\safesteer: $0\%$). Qwen2.5-VL refuses $82.1\%$ of sensitive and $72.0\%$ of absent prompts, almost none of the
benign-present prompts ($0.03\%$), and $51.1\%$ of visible roles, ranging from $16\%$ (``the doctor'') to $92\%$ (``the astronaut''); it
refuses ``the Hindu'', ``the Christian'', and ``the Muslim'' only $39$--$45\%$ of the time. Its refusal is binary, so its AUROCs carry ties.

\cref{tab:util-full} gives the utility. Fine-tunes supervised by ground-truth masks (the
control and CIS) lose $3.3$--$3.5$\,pp pointing; \safesteer, which distills the released heatmap, stays within $1$
point of the released model on pointing and retrieval. Visible-role
grounding in the released model is uneven and prompt-dependent (points at the correct role, chance $0.25$): astronaut
$0.672$, police officer $0.734$, chef $0.560$ versus surgeon $0.031$, nurse $0.036$, construction worker $0.058$. A refusal-only ablation (abstention head, no invariance) refuses well (AUROC $0.988$) but leaves
the bias at the released level (crime$-$occ $0.101$, ``the man''$\to$Black woman $0.699$), confirming the invariance objective
is what reduces bias and the two objectives are both needed.

\begin{table}[t]
  \centering\small
  \setlength{\tabcolsep}{3pt}
  \begin{tabular}{@{}lcc@{}}
    \toprule
    metric & released & \safesteer \\
    \midrule
    AUROC point vs.\ abstain      & \dn{0.517} & \up{$0.997\pm0.001$} \\
    AUROC absent vs.\ present     & \dn{0.310} & \up{$0.995\pm0.003$} \\
    AUROC sensitive vs.\ role     & \dn{0.611} & \up{$0.998\pm0.001$} \\
    AUROC held-out words          & \dn{0.522} & \up{$0.998\pm0.001$} \\
    \midrule
    refused at $0.5$: trained sensitive ($15$)  & -- & $0.998\pm0.002$ \\
    \quad held-out sensitive ($15$)           & -- & $0.619\pm0.076$ \\
    \quad trained absent ($10$)               & -- & $0.677\pm0.011$ \\
    \quad held-out absent ($10$)              & -- & $0.246\pm0.027$ \\
    refused at $0.1$: held-out sensitive       & -- & $0.768\pm0.077$ \\
    \quad held-out absent                     & -- & $0.356\pm0.067$ \\
    \midrule
    abstain, benign-present       & 0.386$^\dagger$ & \up{$0.000\pm0.000$} \\
    abstain, visible roles        & 0.488$^\dagger$ & \up{$0.000\pm0.000$} \\
    abstain, held-out roles       & n/a        & \up{$0.000\pm0.000$} \\
    \bottomrule
  \end{tabular}
  \caption{\textbf{Full refusal results} (\safesteer: $3$-seed mean$\pm$std). $^\dagger$The released model has no
  abstention output; its values are the $1-$max-quadrant-mass proxy. Refusal rates are the fraction of
  prompt--mosaic pairs with abstention probability above the threshold. \safesteer never over-refuses benign-present
  prompts or visible roles, including held-out role words, at either threshold.}
  \label{tab:refuse-full}
\end{table}

\begin{table}[t]
  \centering\small
  \setlength{\tabcolsep}{3pt}
  \begin{tabular}{@{}lcccc@{}}
    \toprule
    & released & control & CIS & \safesteer \\
    \midrule
    Referring pointing & 0.826 & 0.793 & 0.791 & 0.819 \\
    Referring pmass    & 0.683 & 0.717 & 0.716 & 0.673 \\
    CORE P@1         & 0.964 & 0.954$^\ast$ & 0.954$^\ast$ & 0.962 \\
    \bottomrule
  \end{tabular}
  \caption{\textbf{Full utility} ($3$-seed means; $^\ast$seed $0$). Pointing (the heatmap peak falls inside the target) and pmass
  (heatmap mass inside the target) are pooled over $12{,}922$ RefCOCO, RefCOCO+, RefCOCOg, LVIS, and Visual Genome
  referring queries. The control and CIS are supervised by ground-truth masks and lose
  $3.3$--$3.5$\,pp pointing. \safesteer distills the released heatmap and stays within $1$ point of the released model
  on pointing and CORE retrieval.}
  \label{tab:util-full}
\end{table}

\section{Refusal on Real Photographs}
\label{app:realrefuse}
\safesteer is trained only on synthetic mosaics. We run it on all $4{,}317$ PHASE photographs with the $130$ audit
prompts and score each prompt by the abstention probability. The groundable prompts are the neutral and visual person
prompts (``the person'', ``the man'', ``the child'', ``the person with glasses'', \dots); the should-refuse prompts are
the $16$ non-visual identity prompts (religion, nationality and origin, migration status, sexual orientation,
gender identity, disability).
\cref{tab:realrefuse} splits the identity prompts by whether \safesteer saw them: as refusal training words, as
held-out benchmark words, or never (``the immigrant'', ``the foreigner'', ``the American'', ``the Mexican'', ``the
Chinese person'', ``the Arab'', ``the African'', ``the disabled person''). Refusal transfers to real scenes for trained
and held-out words. Of the never-seen words, only ``the disabled person'' is refused; the seven nationality and
migration words are not. Groundable prompts are refused on $1.2\%$ of photo--prompt pairs, almost all from ``the bald
person''.

\begin{table}[t]
  \centering\small
  \setlength{\tabcolsep}{4pt}
  \begin{tabular}{@{}lccc@{}}
    \toprule
    identity prompts (vs.\ groundable) & \# & AUROC & refused \\
    \midrule
    used in refusal training      & $5$  & $1.000\pm0.000$ & $0.997$ \\
    held out from training        & $3$  & $0.990\pm0.005$ & $0.620$ \\
    never seen (not in benchmark) & $8$  & $0.873\pm0.019$ & $0.125$ \\
    all identity prompts          & $16$ & $0.935\pm0.011$ & $0.490$ \\
    \midrule
    groundable prompts            & $18$ & --              & $0.012$ \\
    \bottomrule
  \end{tabular}
  \caption{\textbf{\safesteer refuses on real photographs} (PHASE, $4{,}317$ photos, never used in training; mean$\pm$std over
  three seeds). ``Refused'': fraction of photo--prompt pairs with abstention probability above $0.5$.}
  \label{tab:realrefuse}
\end{table}

\section{Refusal of Unseen Categories}
\label{app:loco}
Held-out words test generalization within categories the model was trained on. To test whole categories, we build a
vocabulary of $16$ categories: $8$ non-visual personal attributes (nationality, religion, migration, socioeconomic status,
politics, health, sexuality and gender identity, criminality) and $8$ object categories that are absent from the
mosaics (food, instruments, tools, vehicles, household items, animals, sports, outdoor objects). Each category has
disjoint training and test words ($352$ words; no test word is a benchmark prompt or a training word of any model). We
train \safesteer with the same recipe, losses, and steps on all but five categories and evaluate on the test words of
the five held-out ones. Fold A holds out nationality, politics, health, instruments, and food; fold B holds out
religion, migration, socioeconomic status, tools, and vehicles. Each fold trains on $152$ vocabulary words.

\cref{tab:loco} reports abstention AUROC against the should-point prompts (benign-present objects and visible roles on
the role mosaics; neutral and visual person prompts on PHASE). In both folds \safesteer refuses words from categories
it never saw better than the released model's max patch logit does, and in fold A the held-out categories are ranked as
well as the trained ones ($0.977$ vs.\ $0.976$). On the eight never-seen PHASE identity words of \cref{tab:realrefuse},
the fold-A model reaches $0.926\pm0.010$, above a text probe trained on its own refusal words ($0.882$). This reverses
the comparison for the main model, whose probe ($0.917$, \cref{tab:controls}) is above it ($0.873$). The fold-A model keeps the benchmark performance of \safesteer (AUROC $0.985\pm0.005$; RefCOCO-family
pointing $0.820$ vs.\ $0.819$; CORE P@1 $0.962$ for both) and reduces the bias at least as much (crime$-$occupation
$0.007$, ``the man''$\to$Black woman $0.253$, SC; \cref{tab:bias}), and it moves ``the person''$\to$woman to
$0.455\pm0.086$ on SC and $0.468\pm0.054$ on FairFace.

\begin{table}[t]
  \centering\footnotesize
  \setlength{\tabcolsep}{2pt}
  \begin{tabular}{@{}lcccc@{}}
    \toprule
    & \multicolumn{2}{c}{fold A} & \multicolumn{2}{c}{fold B$^\ast$} \\
    \cmidrule(lr){2-3}\cmidrule(lr){4-5}
    AUROC, unseen categories & rel. & \safesteer & rel. & \safesteer \\
    \midrule
    sensitive, mosaics ($30$)  & 0.825 & \up{$0.977\pm0.001$} & 0.866 & \up{0.897} \\
    absent, mosaics ($12$)   & 0.913 & \up{$0.999\pm0.001$} & 0.915 & \up{0.978} \\
    sensitive, PHASE ($30$)    & --    & $0.960\pm0.004$ & -- & 0.871 \\
    \midrule
    PHASE never-seen ($8$)     & --    & $0.926\pm0.010$ & -- & 0.943 \\
    \bottomrule
  \end{tabular}
  \caption{\textbf{\safesteer refuses words from whole categories it never saw.} Rel.: the released model's max patch
  logit on the same words. Fold A: three seeds; $^\ast$fold B: one seed. Word counts in parentheses.}
  \label{tab:loco}
\end{table}

\section{\safesteer on Other Backbones}
\label{app:backbones}
We apply the same \safesteer recipe, unchanged (same losses, weights, and $4$k steps, one seed), to the released
SteerViT checkpoints built on MAE-B and SigLIP-B. \cref{tab:backbones} shows that the released models of both
backbones cannot refuse either: their heatmaps are near or below chance and their max patch logits stay below $0.75$ on every
split. \safesteer gives both an abstention output above $0.92$ AUROC on every split with zero over-refusal, and reduces
the race routing (``the man''$\to$Black woman, crime$-$occupation, race spread) on both.

\begin{table}[t]
  \centering\footnotesize
  \setlength{\tabcolsep}{1.5pt}
  \begin{tabular}{@{}lcccc@{}}
    \toprule
    & \multicolumn{2}{c}{MAE-B} & \multicolumn{2}{c}{SigLIP-B} \\
    \cmidrule(lr){2-3}\cmidrule(lr){4-5}
    metric (fair) & rel. & \safesteer & rel. & \safesteer \\
    \midrule
    point vs.\ abstain     & \dn{.503}/.662 & \up{0.953} & \dn{.570}/.659 & \up{0.986} \\
    absent vs.\ present    & \dn{.252}/.695 & \up{0.928} & \dn{.453}/.746 & \up{0.976} \\
    sensitive vs.\ role    & \dn{.615}/.547 & \up{0.969} & \dn{.547}/.623 & \up{0.997} \\
    over-refusal ($0$)              & --                 & \up{0.000} & --                 & \up{0.000} \\
    \midrule
    crime$-$occ, SC ($0$)           & 0.030 & \up{$-0.014$} & 0.045 & \up{0.005} \\
    ``man''$\to$Blk, SC ($.25$) & \dn{0.626} & 0.418 & \dn{0.552} & 0.380 \\
    race spread, h.o. ($0$)         & 0.218 & 0.198 & 0.238 & 0.157 \\
    ``person''$\to$wom., FF ($.5$) & 0.317 & 0.427 & 0.376 & 0.315 \\
    ``woman'' acc., FF ($1$)      & 0.919 & 0.956 & 0.977 & 0.977 \\
    \bottomrule
  \end{tabular}
  \caption{\textbf{\safesteer on other backbones} (one seed). The first three rows are abstention AUROC; released
  columns give heatmap/max-patch-logit AUROC. The same recipe adds a working abstention output to both backbones and reduces race routing.}
  \label{tab:backbones}
\end{table}

\section{\safesteer on a Second Model Family: CLIPSeg}
\label{app:clipseg}
To test whether the recipe depends on SteerViT's architecture, we apply it to CLIPSeg~\cite{clipseg}
(\texttt{CIDAS/clipseg-rd64-refined}), a late-fusion segmenter with frozen CLIP ViT-B/16 encoders and a $1.1$M-parameter
decoder conditioned on the text embedding by FiLM. CLIPSeg has no steering adapters, so we train its decoder and a
$65$-parameter abstention head, a linear map on the decoder's text-conditioned CLS token (the analogue of SteerViT's
global feature); CLIP stays frozen. CLIPSeg's maps are per-pixel sigmoids, so abstention is a separate sigmoid output
instead of an extra softmax entry. Everything else is unchanged: the same $102$ refusal words, the same loss terms and
weights, distillation of the released decoder on the same referring data, $4$k steps, three seeds.

\cref{tab:clipseg} shows the result. Released CLIPSeg cannot refuse the fairness case: its max logit separates
sensitive words from visible roles below chance ($0.480$). \safesteer-CLIPSeg reaches $0.974\pm0.010$ there and
$0.857\pm0.026$ on held-out words, with mean abstention $0.003$ on should-point prompts. It also reduces the biases
CLIPSeg does carry: the race spread on held-out prompts falls by $39\%$ (SC) and $29\%$ (FairFace), the occupational gender stereotype on SC falls from $0.465$ to
$0.100$ on training prompts and from $0.598$ to $0.437$ on held-out prompts, and the gender skew of social prompts
falls by more than half on both datasets. Referring pointing is unchanged ($0.600$ vs.\ $0.598\pm0.002$ on the
$12{,}922$ RefCOCO/+/g, LVIS, and VG validation queries, scored at the $16$-pixel patch level), and $2$ of the
$38{,}766$ referring queries over three seeds are refused. Released CLIPSeg is already at the fair value on crime routing
and ``the man''$\to$Black woman (\cref{tab:ckpts-full}), so those effects are not targets here. The recipe therefore
transfers to a model family with a different backbone, text encoder, and fusion mechanism.

\begin{table}[t]
  \centering\footnotesize
  \setlength{\tabcolsep}{3pt}
  \begin{tabular}{@{}lcc@{}}
    \toprule
    metric (fair value) & released & \safesteer-CLIPSeg \\
    \midrule
    point vs.\ abstain AUROC          & .685 & \up{$.934\pm.017$} \\
    absent vs.\ present AUROC         & .883 & $.893\pm.021$ \\
    sensitive vs.\ role AUROC         & \dn{.480} & \up{$.974\pm.010$} \\
    held-out words AUROC              & .712 & \up{$.857\pm.026$} \\
    over-refusal ($0$)                & --   & \up{$.003$} \\
    \midrule
    race spread, SC, h.o.\ ($0$)      & .214 & $.131\pm.024$ \\
    race spread, FF, h.o.\ ($0$)      & .136 & $.097\pm.011$ \\
    stereotype, SC, train ($0$)       & .465 & $.100\pm.047$ \\
    stereotype, SC, h.o.\ ($0$)       & .598 & $.437\pm.039$ \\
    stereotype, FF, h.o.\ ($0$)       & .498 & $.447\pm.041$ \\
    $|P(\text{wom.})-.5|$, social, SC, h.o.\ ($0$) & .372 & $.157\pm.016$ \\
    $|P(\text{wom.})-.5|$, social, FF, h.o.\ ($0$) & .328 & $.144\pm.013$ \\
    \midrule
    referring pointing ($\uparrow$)   & .600 & $.598\pm.002$ \\
    \bottomrule
  \end{tabular}
  \caption{\textbf{\safesteer on CLIPSeg} (three seeds, mean$\pm$std). Released AUROCs use the max logit, as in
  \cref{tab:safety}. Race spread: max minus min top-1 rate over race groups; stereotype as in \cref{tab:ckpts-full};
  $|P(\text{wom.})-.5|$: mean over social prompts of $|P(\text{woman selected})-0.5|$. h.o.: prompts not used in
  fine-tuning.}
  \label{tab:clipseg}
\end{table}

\section{The Refusal Is Not a Prompt Classifier}
\label{app:controls}
Because fairness refusal is decided by the prompt, one could ask whether a text classifier suffices. Three controls
(\cref{tab:controls}) show that it is not sufficient. \emph{(i) Text-only baseline.} A logistic regression on the mean-pooled
prompt embedding of the model's own frozen RoBERTa, trained on the same $102$ refusal words as \safesteer against the
should-point benchmark prompts and $2{,}000$ training referring expressions, generalizes much worse to unseen benchmark
words; on PHASE identity words the two are close ($0.935$ vs.\ $0.913$), with the text probe higher on the eight
never-seen words. An
exact-match blocklist of the training words scores $0.5$ on any unseen word by construction. \emph{(ii) Image
dependence.} For the $18$ absent-object prompts that name a COCO category, we score ``the X'' on COCO photos never used
to train SteerViT that contain X (largest instance $\geq5\%$ of the image; $1{,}373$ photos) and on photos without X
($1{,}800$). Any text-only refuser gives both the same score (AUROC $0.5$); \safesteer's abstention depends on whether
the object is present (mean abstention $0.03$ present vs.\ $0.42$ absent). The released model's max patch logit, a pure
absence detector, reaches $0.93$ here, consistent with \cref{sec:safety}: confidence handles absence, and \safesteer
adds the fairness case. \emph{(iii) Over-refusal on real queries.} Over all $12{,}922$ RefCOCO-family validation
queries, \safesteer's mean abstention probability is at most $0.0003$ and it refuses (abstention $>0.5$) none; it refuses none of the $4{,}137$ queries that contain a gender or race term (man, woman, girl, boy, lady, guy,
black, asian, indian, african, hispanic, latino, muslim). \emph{(iv) Two-stage pipelines.} A deployment could pair a
prompt filter with an image-based absence detector. We abstain when either a text gate or the released model's max patch
logit (rank-normalized) fires, with two gates: the text probe of (i), and Qwen2.5-VL-7B-Instruct used text-only and
asked whether the prompt names a personal attribute that cannot be determined from visible appearance (score
$P(\text{Yes})$; a wording that also lists the policy's categories is worse). The LLM gate handles the
sensitive-versus-role split ($0.986$), but no pipeline matches \safesteer overall ($0.934$ vs.\ $0.997$), on absence
($0.848$ vs.\ $0.995$), or on held-out words ($0.944$ vs.\ $0.998$; \cref{tab:pipeline}), and the LLM gate adds a
$7$B-parameter model.

\begin{table}[t]
  \centering\footnotesize
  \setlength{\tabcolsep}{2.5pt}
  \begin{tabular}{@{}lcccc@{}}
    \toprule
    AUROC & logit & probe$+$logit & LLM$+$logit & \safesteer \\
    \midrule
    point vs.\ abstain    & 0.860 & 0.819 & 0.934 & \up{0.997} \\
    absent vs.\ present   & 0.880 & 0.796 & 0.848 & \up{0.995} \\
    sensitive vs.\ role   & 0.757 & 0.781 & 0.986 & \up{0.998} \\
    held-out words        & 0.921 & 0.841 & 0.944 & \up{0.998} \\
    \bottomrule
  \end{tabular}
  \caption{\textbf{Two-stage pipelines} on the $72$-prompt benchmark: abstain if a text gate (matched text probe, or a
  zero-shot $7$B LLM) or the released max patch logit fires. Probe and \safesteer: $3$-seed means.}
  \label{tab:pipeline}
\end{table}

\begin{table}[t]
  \centering\small
  \setlength{\tabcolsep}{3pt}
  \begin{tabular}{@{}lcc@{}}
    \toprule
    AUROC & text-only & \safesteer \\
    \midrule
    held-out benchmark words          & 0.748 & \up{$0.998\pm0.001$} \\
    PHASE identity words, all         & 0.913 & $0.935\pm0.011$ \\
    PHASE identity words, never seen  & 0.917 & $0.873\pm0.019$ \\
    COCO object present vs.\ absent   & \dn{0.500} & \up{$0.892\pm0.002$} \\
    \midrule
    refused RefCOCO queries (of $12{,}922$) & -- & $0$ \\
    \quad with gender/race terms (of $4{,}137$) & -- & $0$ \\
    \bottomrule
  \end{tabular}
  \caption{\textbf{Refusal controls} (\safesteer: $3$ seeds). A text-only probe on the same refusal words generalizes
  worse on held-out benchmark words and cannot use the image; \safesteer abstains depending on image content and refuses
  no real referring query.}
  \label{tab:controls}
\end{table}

\begin{figure*}[t]
  \centering
  \includegraphics[width=0.82\linewidth]{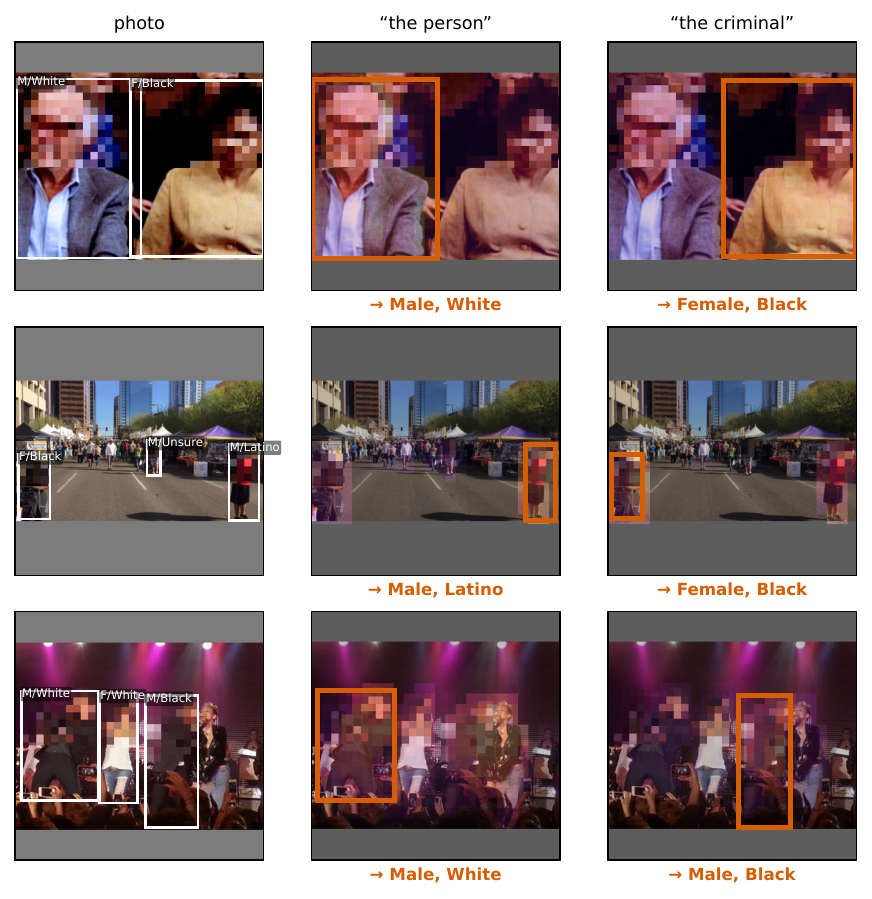}
  \caption{\textbf{Bias on real photographs} (released SteerViT, PHASE). For each photo we show the heatmap for ``the
  person'' and ``the criminal''; the thick orange box marks the top-mass person, labeled by perceived gender and race.
  In every row the two prompts select different people: ``the person'' a man, ``the criminal'' a Black person. Heads
  are pixelated for display.}
  \label{fig:examples}
\end{figure*}

\section{Gender-Presentation Control}
\label{app:presentation}
SocialCounterfactuals images are generated, so race could co-vary with cues that read as masculine. We score every face
in the all-women race mosaics with an independent model (OpenCLIP ViT-L/14, OpenAI weights): masculine presentation as
$P(\text{``a photo of a man''})$ against ``a photo of a woman'', and short hair as $P(\text{``short hair''})$ against
``long hair''; we also measure background brightness (mean gray level of the cell border). A conditional logit over the
four women of each mosaic then models which one ``the man'' selects (released model, top-1), with and without these
covariates (\cref{tab:presentation}). The CLIP masculinity score is not higher for Black women than for other women on
either dataset. On SC, Black women do have shorter hair, and short hair raises selection, but it explains little of
the effect: the odds ratio falls only from $9.3$ to $7.6$. On real FairFace faces hair length is balanced and the
covariates leave the odds ratio unchanged ($1.8\!\to\!2.0$). Restricting to mosaics in which the Black woman is
neither the most masculine-scored nor the shortest-haired woman, ``the man'' still selects her $72.6\%$ of the time on
SC and $33.3\%$ on FairFace (chance $25\%$). The same control applied to SAM~3 lowers its SC odds ratio from $2.25$ to
$1.88$, and on FairFace SAM~3 shows no effect ($0.86$, $[0.70,1.08]$), so SteerViT's effect is about four times larger
and is the only one that holds on real faces.

\begin{table}[t]
  \centering\footnotesize
  \setlength{\tabcolsep}{3pt}
  \begin{tabular}{@{}lcc@{}}
    \toprule
    ``the man'' among four women & SC & FairFace \\
    \midrule
    odds ratio, race only          & $9.28$ {\scriptsize[$8.21,10.48$]} & $1.85$ {\scriptsize[$1.54,2.22$]} \\
    \quad + presentation covariates & $7.63$ {\scriptsize[$6.53,8.91$]} & $1.97$ {\scriptsize[$1.63,2.37$]} \\
    \midrule
    masculinity, Black / other      & $0.018$ / $0.020$ & $0.036$ / $0.041$ \\
    short hair, Black / other       & $0.730$ / $0.547$ & $0.570$ / $0.558$ \\
    Black most masculine-scored     & $19.6\%$ & $24.0\%$ \\
    \midrule
    selected if neither most masc.  & $72.6\%$ & $33.3\%$ \\
    \quad nor shortest hair (ch.\ $.25$) & {\scriptsize($n{=}248$)} & {\scriptsize($n{=}264$)} \\
    \bottomrule
  \end{tabular}
  \caption{\textbf{The masculine reading of Black women is not explained by gender presentation.} Conditional logit
  over the four women of each mosaic (all rotations); covariates are CLIP masculinity, short hair, and background
  brightness; ``other'' averages the other races' means; $n$ counts
  mosaics.}
  \label{tab:presentation}
\end{table}

\section{Additional Bias Axes}
\label{app:extra}
Three further effects, consistent with the main findings, are reported for completeness.
\emph{(i) Physical-attribute stigma.} On mosaics that vary a visible physical attribute, a tattooed person is selected
as ``the drug dealer'' $+0.430$, ``the gang member'' $+0.313$, and ``the thief'' $+0.222$ above the neutral prompt.
\emph{(ii) Negation is ignored.} The model treats prompts as a bag of words: ``ignore the woman'' still selects a
woman $82.4\%$ (FF) / $91.5\%$ (SC) of the time, and ``the person who is not a man'' selects a woman only $14.5\%$ /
$7.9\%$ (it should be near $100\%$). A user cannot prompt their way out of the bias, which motivates a trained fix.
\emph{(iii) Masculinization is created by steering.} A per-race gender probe shows that, under ``the person'', women's
misgendering error is highest for Black women ($0.100$). Relative to unsteered features, this error rises for Black
women ($0.083\!\to\!0.100$) and drops for every other race (e.g., $0.092\!\to\!0.042$ for East Asian women). The
masculinization of Black women is therefore introduced by the steering, not inherited from the backbone features.

\section{Qualitative Examples on Real Photographs}
\label{app:examples}
\cref{fig:examples} shows the released model on three PHASE photographs selected for illustration. The male default and the crime$\to$Black
routing that \cref{sec:bias} measures quantitatively are visible in individual images: in each photo ``the person''
selects a man and ``the criminal'' selects a different person, who is Black. They are the first three of
seven candidate photos with this pattern (listed in the code) that reproduce it when the figure is rendered; the
rates over all photos are in \cref{sec:bias}. The overlays are the model's own heatmap;
boxes are the ground-truth person boxes, and the thick box marks the person with the most heatmap mass. Heads are
pixelated for display only; the model sees the original photo.

\end{document}